\documentclass[9pt,twocolumn,twoside]{osajnl}
\usepackage{times}
\usepackage{amssymb}
\usepackage{multirow}
\usepackage{mathtools}
\usepackage{subfloat}
\usepackage{bbm}
\usepackage{algorithm}
\usepackage{algpseudocode}
\usepackage{pbox}
\usepackage{placeins}
\definecolor{grey}{rgb}{0.95,0.95,0.95}
\journal{ol} 
\setboolean{shortarticle}{false}

\usepackage{times}
\usepackage{graphicx}
\usepackage{units}
\usepackage{url}

\usepackage{amsmath}

\usepackage[printonlyused]{acronym}
\newacro{DoF}{degree of freedom}
\newacroplural{DoF}{degrees of freedom}
\acrodefplural{DoF}[DoFs]{degrees of freedom}
\newacro{iLQR}{Iterative Linear-Quadratic Regulator}
\newacro{CT}{Control Toolbox}
\newacro{FoV}{field of view}
\newacro{EOM}{equations of motion}
\newacro{OC}{Optimal Control}
\newacro{LQR}{linear-quadratic regulator}
\newacro{PD}{proportional derivative}
\newacro{MPC}{Model Predictive Control}
\newacro{LQ}{linear quadratic}
\newacro{LQOC}{Linear-Quadratic Optimal Control}
\newacro{TO}{Trajectory Optimization}
\newacro{DDP}{Differential Dynamic Programming}
\newacro{COM}{center of mass}
\newacro{COP}{center of pressure}
\newacro{NLP}{nonlinear program}
\newacro{MLP}{Multilayer Perceptron}
\newacro{SLQ}{Sequential Linear-Quadratic}
\newacro{HAA}{hip abduction adduction}
\newacro{AD}{automatic differentiation}
\newacro{HJB}{Hamilton–Jacobi–Bellman}
\newacro{BC}{Behavioral Cloning}
\newacro{IRL}{Inverse Reinforcement Learning}
\newacro{IL}{Imitation Learning}
\newacro{RL}{Reinforcement Learning}
\newacro{MEN}{mixture-of-experts network}
\newacro{MILE}{mixture of implicitly localized experts}
\newacro{MELE}{mixture of explicitly localized experts}
\newacro{DL}{Deep Learning}
\newacro{WBC}{whole-body controller}
\newacro{MDP}{Markov Decision Process}
\newacro{RNN}{Recurrent neural network}
\newacro{PPO}{Proximal Policy Optimization}
\newacro{N-P3O}{Normalized Penalized Proximal Policy Optimization}
\newacro{EKF}{Extended Kalman Filter}
\newacro{HSV}{Hue-Saturation-Value}
\newacro{EE}{End-Effector}
\newacro{FOV}{field of view}
\newacro{CoT}{cost of transport}
\newacro{GT}{ground truth}
\newacro{QDD}{quasi direct drive}
\newacro{SLAM}{simultaneous localization and mapping}
\newacroplural{QDD}{quasi direct drive}
\acrodefplural{QDD}[QDDs]{quasi direct drives}
\newacro{SEA}{series elastic actuator}
\newacroplural{SEA}{series elastic actuator}
\acrodefplural{SEA}[SEAs]{series elastic actuators}

\ifx\isReview\undefined
    
\else
    \usepackage{xcolor}
    
\fi

\title{Learning coordinated badminton skills for legged manipulators} 

\author{
Yuntao Ma$^{1\ast}$,
Andrei Cramariuc$^{1}$,
Farbod Farshidian$^{2}$,
Marco Hutter$^{1}$
\\
\normalsize{$^{1}$Robotic Systems Lab, ETH Zurich, 8092 Zurich, Switzerland.} \\
\normalsize{$^{2}$Currently at Robotics and AI Institute, 145 Broadway, Cambridge MA, USA.} \\
\normalsize{$^\ast$Corresponding author. Email: mayuntao94@gmail.com} \\
}

\dates{Accepted on 29 April, 2025}

\begin{abstract}
Coordinating the motion between lower and upper limbs and aligning limb control with perception are substantial challenges in robotics, particularly in dynamic environments. To this end, we introduce an approach for enabling legged mobile manipulators to play badminton, a task that requires precise coordination of perception, locomotion, and arm swinging. We propose a unified reinforcement learning-based control policy for whole-body visuomotor skills involving all degrees of freedom to achieve effective shuttlecock tracking and striking. This policy is informed by a perception noise model that utilizes real-world camera data, allowing for consistent perception error levels between simulation and deployment and encouraging learned active perception behaviors. Our method includes a shuttlecock prediction model, constrained reinforcement learning for robust motion control, and integrated system identification techniques to enhance deployment readiness. Extensive experimental results in a variety of environments validate the robot's capability to predict shuttlecock trajectories, navigate the service area effectively, and execute precise strikes against human players, demonstrating the feasibility of using legged mobile manipulators in complex and dynamic sports scenarios.
\end{abstract}

\begin{document} 
\maketitle

\begin{figure*}[htp]
    \centering
    \includegraphics[width=0.8\textwidth]{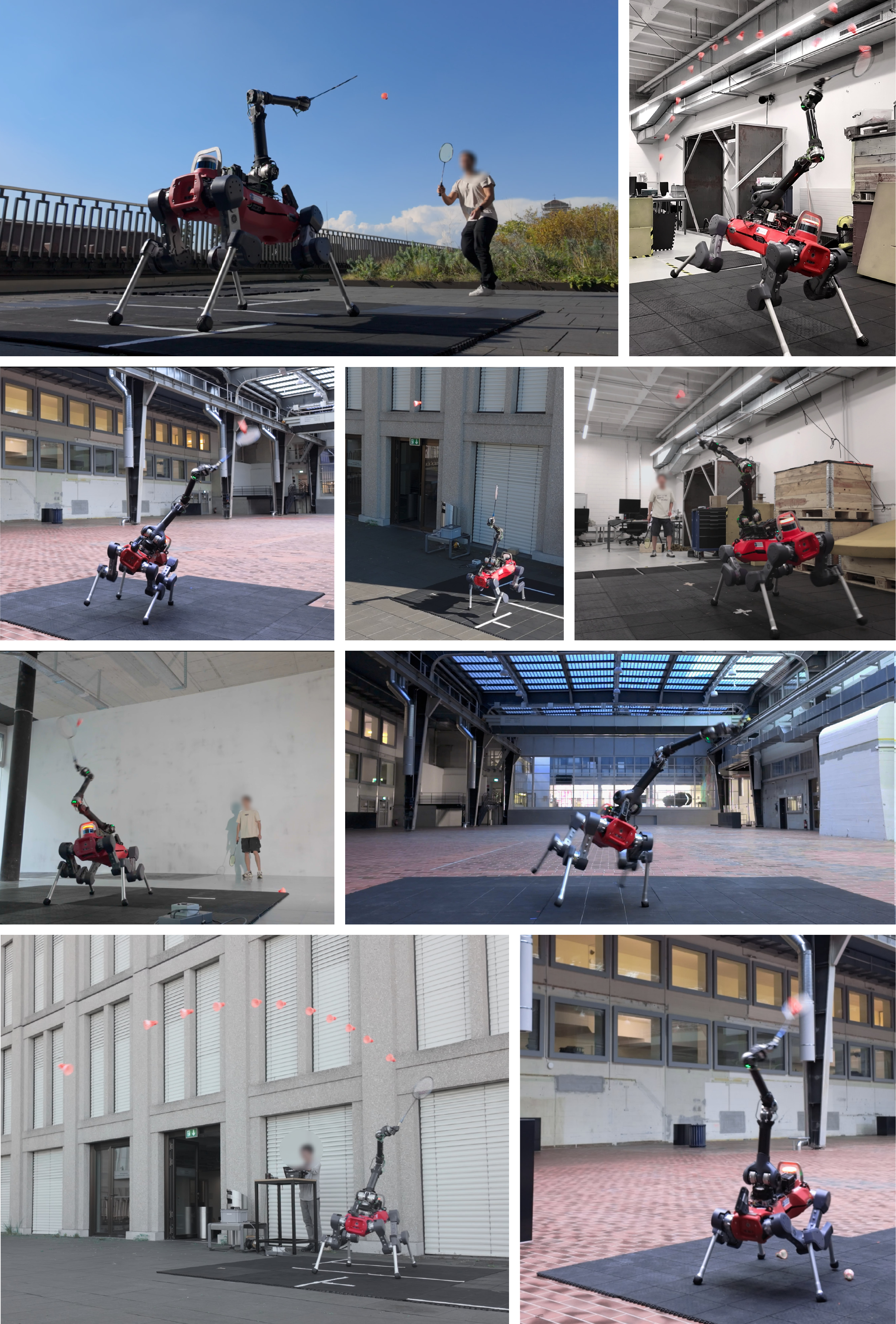}
    \caption{\textbf{Deployment of the badminton control policy on our legged manipulator.} Our system operates entirely on onboard perception and computation for shuttlecock detection, trajectory prediction, and limb control. It has been tested in various environments, including the lab, a historic machine hall, and outdoor settings.}
    \label{fig:photo_collage}
    
    % \vspace{-1.5em}
\end{figure*}

% add badminton shuttlecock trajectory

\section*{Introduction}
Human sport competitions like badminton pose substantial challenges for players due to the complex interplay required between footwork and upper limb movements. Competent players must develop advanced loco-manipulation skills to effectively cover the extensive court area, complemented by precise hand-eye coordination to anticipate and correctly hit the shuttlecock. Players must predict the shuttlecock's most likely incoming trajectory to synchronize the timing, location, angle, and velocity for strikes that achieve a successful return.

This complex interplay of perception, locomotion, and manipulation makes such sports applications a formidable challenge for developing advanced unified skills for legged mobile manipulation systems as it tackles deficiencies in current paradigms for controllers and hardware. The main challenge for such robots involves balancing rapid and responsive locomotion with accurate arm movements. Although the robot's high number of \acp{DoF} in principle allows for agile movements, realizing such potential in practice depends heavily on the control algorithms.

This balancing act is further complicated by the limitations of commercial onboard camera systems, which often must compromise between frame rate, angular resolution, \ac{FOV}, and transmission delay, in contrast to the human eye, which has much better motion stabilization, adjustable focus and more sophisticated information processing. To attain similar performance, digital cameras on robots require a perception-aware controller that moves the camera using smooth motions while keeping the target in the \ac{FOV}. 

Various methods have been used for athletic robot control in prior works. Model-based control methods have been utilized for executing complex maneuvers such as front/backflips~\cite{raibert1986legged,le2024fast,katz2019mini} and object throwing~\cite{liu2024tube,chiu2022collision}. These techniques either require model simplifications or only allow local feedback control around pre-optimized trajectories. Recently, noteworthy progress in \ac{RL}-based robot running~\cite{crowley2023optimizing, ji2022concurrent, margolis2024rapid,he2024agile} and parkour~\cite{hoeller2024anymal, cheng2023extreme, zhuang2023robot, caluwaerts2023barkour} has demonstrated the potential of learning-based methods to advance robot agility. However, these developments are primarily restricted to locomotion in static scenes. Moreover, the exploration of pedipulation -- namely employing robots' feet as manipulators -- through \ac{RL} has uncovered promising approaches to enhance robots' applicability in interactive sports~\cite{ji2023dribblebot, haarnoja2024soccer}. Despite its potential, this method is generally characterized by a limited operational reach, potentially curtailing its utility in activities where extensive spatial interaction is essential.

In terms of application, one of the closest sports is table tennis, which has been extensively researched for both accuracy~\cite{d2024achieving, ding2022learning, ma2023reinforcement} and strategy~\cite{abeyruwan2023sim2real, d2024achieving}, primarily using fixed-base or gantry manipulators with external vision systems. In contrast, our work emphasizes whole-body visuomotor skills and relies solely on onboard perception, integrating both legged locomotion and arm swinging — an approach that more closely mirrors human gameplay.

Specifically for legged manipulators, to date, the integration of locomotion and manipulation within the realm of legged manipulation controllers often sees these tasks being treated as distinct entities~\cite{ferrolho2023roloma,zimmermann2021go,ma2022combining,sleiman2021unified,yokoyama2023asc,liu2024visual,dao2024box}. This decoupled approach simplifies the optimization process; however, it imposes substantial constraints on the robot's range of motion and its agility, which are critical factors in dynamic environments. Several recent studies have deviated from this traditional decoupling paradigm, albeit with modifications that either maintain a slow-moving state \cite{fu2023deep} or focus solely on self-manipulation tasks \cite{ma2023learning}, which does not fully exploit the manipulation capabilities of legged robots in more dynamic settings. Conversely, there has been noteworthy progress in achieving dynamic motions through imitation learning. Techniques developed for simulated agents playing tennis \cite{zhang2023vid2player3d} or humanoid robots engaging in generic imitative behaviors \cite{fu2024humanplus,cheng2024expressive} have showcased the potential of using demonstration data to inform robotic control systems while requiring demonstrations from agents of similar morphology. A recent approach involving task-space imitation learning on quadrupedal mobile manipulator platforms has demonstrated dynamic manipulation capabilities with the arms and grippers while not fully leveraging the locomotive potential of the legs in hardware deployments~\cite{ha2024umilegsmakingmanipulation}.

Another major challenge is the trade-off between active perception behavior and agile motion control. Privileged learning has been employed in scenarios where a student policy reconstructs privileged teacher observations based on past interactions and perception history \cite{lee2020learning,miki2022learning,agarwal2023legged}. In this framework, the teacher policy observes privileged information and is not incentivized to learn active perception behaviors, namely, visiting trajectories are informative for decoding the privileged observations. This results in an information gap between the teacher and student policies. Some recent works incorporating perception directly into the RL training loop represent an advancement. For instance, neural rendering techniques utilize a learned perception model where the latent representations can be efficiently integrated within the training process \cite{hoeller2022neural, hoeller2024anymal}. This approach has shown efficacy in structured and static scenes, enabling the potential learning of active exploration behaviors. Furthermore, the emergence of active perception behaviors has been documented in recent research \cite{schwarke2023curiosity, team2021open}. However, the emergent active perception behavior was not quantitatively evaluated in these works.

Additional practical deployment challenges include the constraints on electrical current supply to robot actuators, the necessity for precise system identification for accurate simulation modeling, and the need to mitigate perception and communication delays. These elements collectively pose barriers to developing robots that can play badminton at a level comparable to human players.

To address the complex control challenges encountered in real-scale badminton games, we developed a unified \ac{RL}-based controller trained in simulation, that controls both the base locomotion and the arm manipulative actions. Extending previous work \cite{rudin2022advanced}, this integrated approach leverages all \acp{DoF} of the robot to track target \ac{EE} states at specified points in time, enabling effective responses to the shuttlecock's incoming trajectory. To prepare the control policy for consecutive shuttlecock hits and learn post-swing follow-through behaviors, we implemented multiple swing targets that are \unit[2]{s} apart per learning episode. To ensure that the \ac{MDP} is well-posed for the critic to estimate the value function, we employed an asymmetric actor-critic formulation \cite{pinto2017asymmetricactorcriticimagebased}. We incorporated a parameterized shuttlecock perception model based on real-world camera data in the training loop. This model captures the effect of robot motion on perception quality by accounting for both single-frame object tracking errors and final interception predictions, which reduces the perception sim-to-real gap and allows the robot to learn perception-driven behaviors that are also effective on the hardware. This perception model enabled us to reward the policy based on the final perception error and avoids the need for hard-coded orientation strategies, preserving motion efficiency. As an example, the robot may pitch up to keep the shuttlecock in the camera \ac{FOV} until it needs to pitch down again to swing the racket. The \ac{RL} algorithm balances the trade-off between agile control and accurate shuttlecock perception by optimizing the policy's overall ability to hit the shuttlecock in simulation. Furthermore, the system integrates shuttlecock prediction \cite{cohen2015physics}, constrained \ac{RL}\cite{lee2023evaluation}, state estimation\cite{khattak2020complementary,lynen13robust}, and system identification on the manipulator dynamics\cite{hwangbo2019learning,filip2024,tan2018sim} to facilitate the hardware deployment of the trained control policy.

In this Article, we demonstrate a quadrupedal mobile manipulator that autonomously plays badminton with human opponents using only onboard perception. Through an integrated \ac{RL} approach that coordinates whole-body motion with perception, the robot adapts its gait patterns based on time and distance constraints to track and intercept shuttlecocks with swing velocities up to \unit[12.06]{m s$^{-1}$}. Extensive hardware and simulation experiments validate the system's ability to maintain long rallies in collaborative matches with humans, showcasing emergent active perception behaviors and consistent shuttlecock interception within the court, all while balancing stability with agile arm swings.

\begin{figure}[t]
    \centering
    % Link the image to your desired URL or file
    \href{https://youtu.be/zYuxOVQXVt8}{%
        \includegraphics[width=0.48\textwidth]{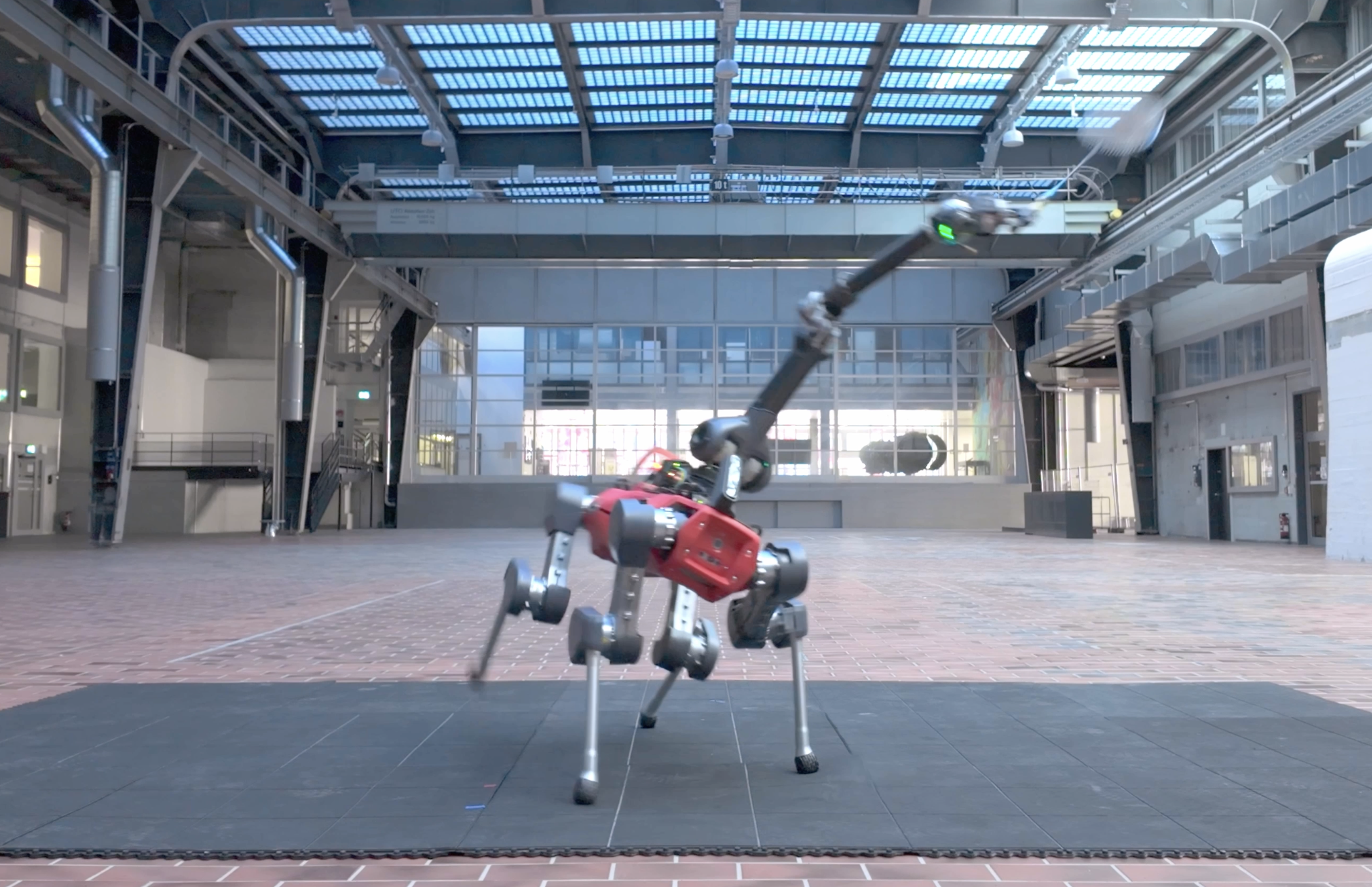}  % Adjust width as needed
    }
    % Use \caption* to create a custom caption without "Figure" prefix
    \caption*{\textbf{Movie 1. Summary of the results and the method.} The video demonstrates our approach for enabling a legged mobile manipulator robot to play badminton through coordinated control. It showcases how our training pipeline produced policies that balance mobility requirements with rapid arm movements for successful shuttlecock returns. The video also highlights our active perception framework, which incorporated real-world camera noise models into reinforcement learning to develop perception-aware behaviors. This allowed the robot to track fast-moving shuttlecocks while traversing the court for interception and return shots. Various experiments with human players illustrated the system's capabilities across different gameplay scenarios.}
\end{figure}

\section*{Results}

Movie 1 summarizes the results of the presented work. Our experiments demonstrated the robot's ability to autonomously track, intercept, and return shuttlecocks during gameplay with human opponents. The system successfully coordinated whole-body movements, adapting its posture and gait patterns based on the shuttlecock's trajectory, while maintaining effective visual tracking. We evaluated the robot's performance in aspects including interception success rates, swing velocity tracking, active perception capabilities, and adaptive locomotion strategies.

\subsection*{System overview}

\begin{figure}[ht!]
    \centering
    \includegraphics[width=0.5\textwidth]{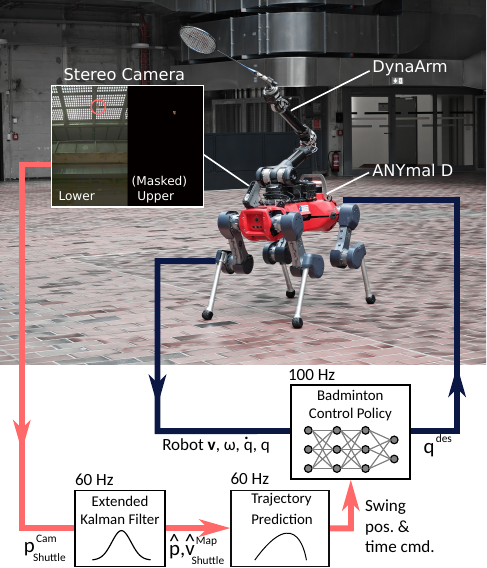}
    \caption{\textbf{System overview. } The legged mobile manipulator consists of a quadrupedal base and a dynamic arm. We additionally mounted a stereo camera with global shutters. The robot controller system receives the shuttle position computed in the camera frame, predicts the interception position, and feeds it to the RL policy along with the robot proprioception observations. The policy controls all 18 robot drives by producing joint commands.}
    \label{fig:overview}
    
    % \vspace{-1.5em}
\end{figure}

The quadrupedal mobile manipulator robot used in this work consisted of the ANYmal-D base \cite{hutter2016anymal} and the DynaArm. The robot was equipped with a ZED X stereo camera with global shutters for shuttlecock perception (Fig. 2A). The badminton racket was oriented at a 45-degree angle with respect to the wrist joint, which proved to be the most effective configuration based on early simulation tests of various orientations.

For deployment, the robot's state estimation operated at a frequency of \unit[400]{Hz}, and the robot control policy updated observations and sent joint position commands at a rate of \unit[100]{Hz}, as illustrated in Fig. 2. The system's perception included shuttle position measurement, state estimation, and trajectory prediction. It ran asynchronously at \unit[60]{Hz} on a Jetson AGX Orin module. Further details are available in the Materials and Method section.

\subsection*{Collaborative game with human players}
Collaborative games were held between the robot and amateur players to validate the system's capability to play real-scale badminton and maintain long rallies, as shown in Movie 1. Throughout the games, ANYmal was able to respond appropriately to the incoming shuttle with various velocity and landing positions, albeit with some failures to return them. The perception module took on average \unit[0.357]{s} after the opponent had hit the shuttlecock to register trajectories for interception. This left on average \unit[0.654]{s} until the shuttle trajectory crossed the target interception height of \unit[1.25]{m} above the robot base height. The fastest hit from the policy was \unit[0.367]{s} after the interception position was computed.

The robot was capable of consecutive hits. Multiple rallies were documented in Movie 1, with a streak of 10 shots in a single rally. The policy also demonstrated the emergent behavior of moving back near the center of the court after each hit, similar to how human players prepare for the next hits.

\subsection*{Badminton motor skills}

\begin{figure*}[ht!]
    \centering
    \includegraphics[width=0.9\textwidth]{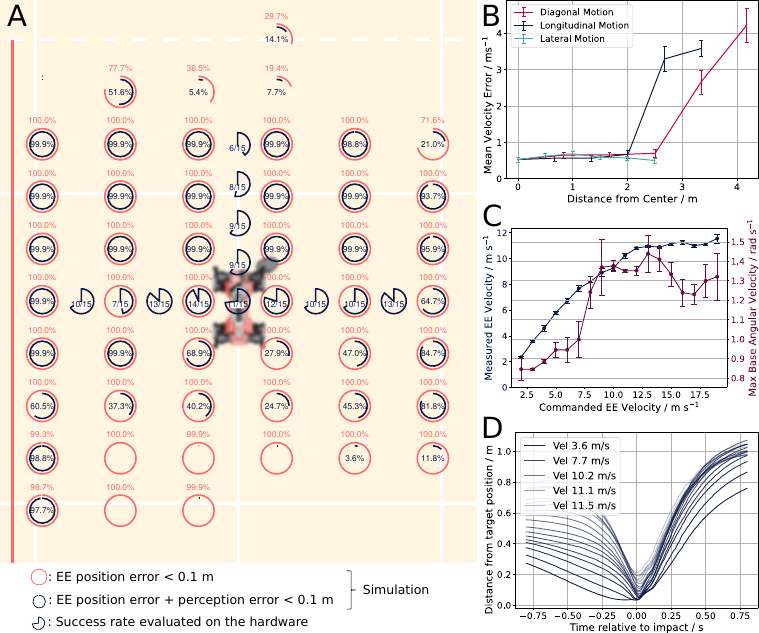}
    \caption{\textbf{Assessing the controlled racket swings. (A)} We quantitatively evaluated the control system's success rate with progressively more difficult conditions in simulation and partly on the hardware. The EE position error condition evaluates the robot's ability to reach EE targets across the court; the additional perception error condition assesses the robot's ability to perform coordinated perception and control. On the hardware, we evaluated the robot's ability to hit the shuttlecock back across the net at intervals of \unit[0.5]{m}. The target and robot's initial positions are drawn to their corresponding position on the badminton court. \textbf{(B)} The EE velocity tracking error at interception increases as the robot attempts to hit shuttlecocks landing further from the court center. The error bars represent the standard deviation of the error across 6600 trials. \textbf{(C)} The EE swing velocity and maximum base angular velocity are plotted against the commanded EE velocity, with the error bar spanning from the minimum to maximum values. The robot is able to reach a maximum executed swing velocity of \unit[12.06]{ms$^{-1}$}, with generally increased base angular velocities at higher EE velocity commands.  \textbf{(D)} The distance between the racket sweet spot and the target impact position over relative time. The end-effector minimizes this distance precisely at the commanded impact time. The evaluations in \textbf{(C-D)} are conducted on the hardware.}
    \label{fig:success_rate}
    
    % \vspace{-1.5em}
\end{figure*}

To evaluate the effectiveness of our control policy, we measured the overall success rate of the robot in hitting shuttlecocks that would land at various distances. Fig. 3A indicates the simulated and hardware success rates in different parts of the badminton court, with the robot always starting at the center. The shuttlecock trajectories in this evaluation followed the same initial velocity, and the initial positions were shifted such that the landing locations were distributed as in the figure. We assessed in simulation the policy's ability to intercept the shuttlecock with increasing levels of difficulty, each level corresponding to a qualitative improvement of the badminton skill.

\subsubsection*{Level I - position tracking}
Given ground truth perception, we evaluated the percentage of hits that reached the interception position within \unit[0.1]{m} - the approximate distance between the racket's center and its edge - at the commanded swing time. In the service area, the simulated results indicated the robot could intercept the incoming shuttlecock with a negligible failure rate.

\subsubsection*{Level II - perception error}
To assess the robot's ability to predict the shuttle trajectory and intercept it successfully, we added the perception error to the success criteria for this level of evaluation. The perception error measures the position difference between the ground truth shuttlecock position in the simulation and the \ac{EKF}-estimated position at the time of the swing. To achieve a small error, the control policy had to command the robot to maintain sight on the shuttlecock for a substantial duration based on the measurement noise model while accurately tracking the racket state at the commanded point in time. This level represents the robot's ability to predict and intercept the shuttlecock successfully. This task became particularly challenging at the borders of the service court and when the shuttlecock landed directly behind the robot, as indicated by the lower success rate in Fig. 3A in these regions. We attributed this difficulty to the robot's rectangular FOV, which synergized well with base tilting maneuvers (shown in Fig. 1) to extend visual tracking of the shuttlecock. However, when the shuttlecock approached from directly overhead or behind, the robot had to pitch directly upwards, making it substantially more challenging to maintain continuous visual contact.

Additionally, we reported the \ac{EE} velocity tracking error at interception when the robot attempted to hit shuttlecocks landing further from the court center along lateral, longitudinal (front-back), and diagonal directions. The velocity tracking accuracy degraded when shuttlecock landings occurred beyond \unit[2.5]{m} from the court center in the lateral and diagonal directions and beyond \unit[2.0]{m} in the longitudinal direction.

We validated the success rate evaluation on the hardware by examining how effectively the robot could hit the shuttlecock back over the net. The hardware evaluation was conducted with the robot facing the net while intercepting shuttlecocks approaching from both lateral and frontal directions. This validation confirmed the effectiveness of our deployed control method. Notably, the robot maintained stability and avoided the arm current consumption constraint throughout the hardware experiments, demonstrating its robustness. Video documentation of this evaluation is available in the supplementary movie S1.

\subsection*{Fast and accurate racket swing}
We further assessed the system's ability to track swing velocity and position commands on hardware with the setup shown in supplementary movie S2. The robot was commanded to swing at varying target velocities to reach position targets at the center of the court at a height of \unit[1.3]{m} above the starting base height. In Fig. 3B, the executed \ac{EE} velocity and maximum base angular velocity were plotted against the commanded velocities. The executed swing velocity generally tracked the commanded velocity below \unit[10]{m s$^{-1}$}, with diminishing accuracy at higher velocities. The robot achieved a peak executed velocity of \unit[12.06]{m s$^{-1}$} when commanded to swing at \unit[19]{m s$^{-1}$}. By comparison, amateur badminton players can reach swing velocities between \unit[20–30]{m s$^{-1}$}, and a recent study on robot table tennis \cite{d2024achieving} reported an average swing velocity of \unit[6.83]{m s$^{-1}$} for their fastest low-level skill. As detailed in the Materials and Methods section, the system operated near its current and joint velocity limits to achieve these commands. Additionally, higher commanded velocities led to increased base angular velocities, indicating a coupling between base attitude control and manipulator swing.

Fig. 3C shows the distance between the racket and the target position around the impact time, with the racket reaching its closest point precisely at the commanded impact moment. At the commanded swing of \unit[12]{m  s$^{-1}$}, the robot executed swing of mean \unit[10.8]{m s$^{-1}$}, with a mean position error of \unit[0.117]{m}, which, in terms of timing, was equivalent to a mere offset of \unit[0.0108]{s} as the racket moved at the target velocity.

\subsection*{Active perception behavior}

\begin{figure*}[ht!]
    \centering
    \includegraphics[width=0.9\textwidth]{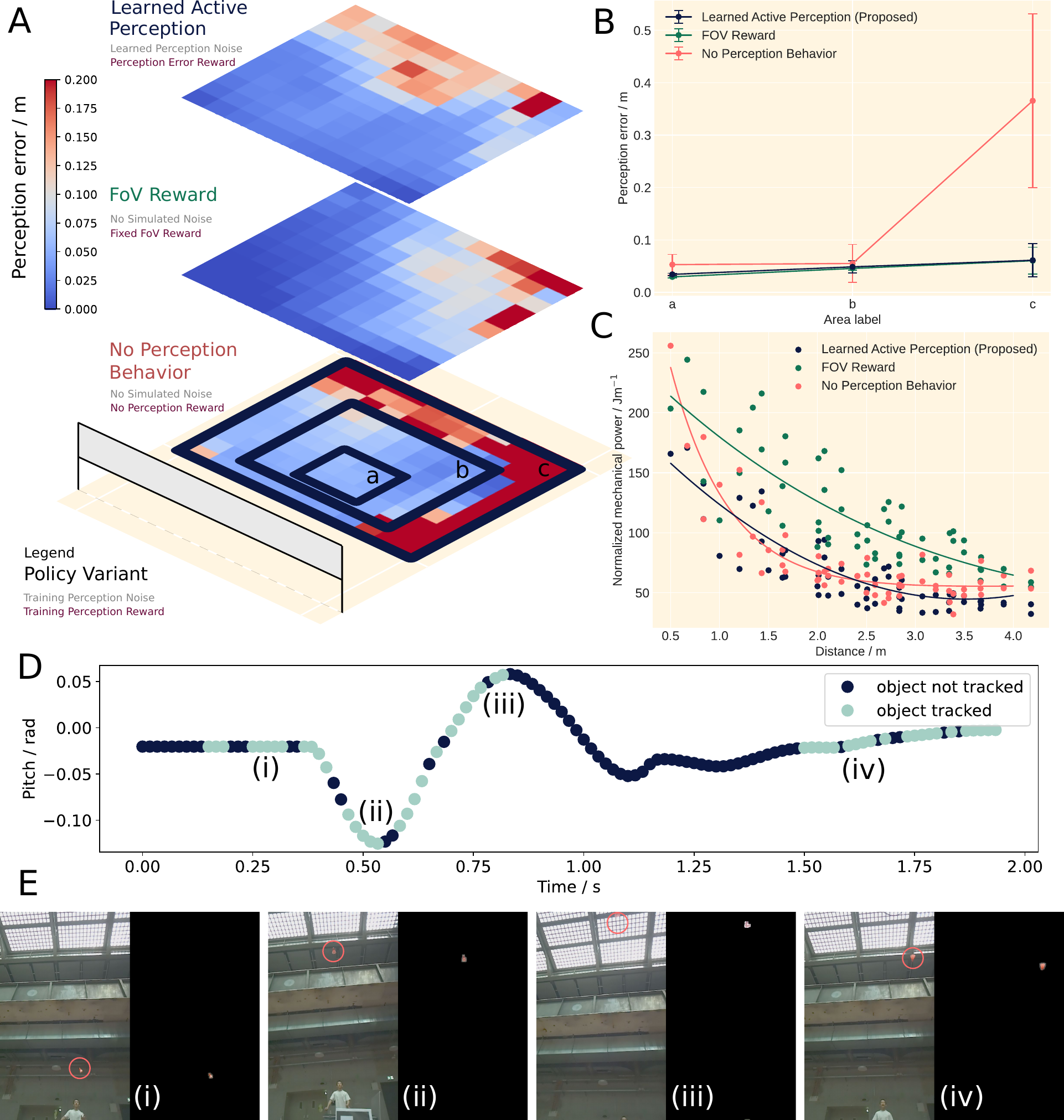}
    \caption{\textbf{Learned Active Perception. (A)} A comparison of the perception error heatmap at the time of desired shuttlecock contacts. The policy trained with ground truth shuttlecock states is not incentivized to actively track the shuttlecock, leading to a larger perception error than the other two methods. The heatmaps are overlaid with the badminton court. \textbf{(B)} The perception error statistics collected based on different areas in the badminton court. The line graph shows the mean error, and the error bars represent the standard deviation, computed from 1650 trajectories per cell in each region. \textbf{(C)} The mechanical power required to execute racket swings, normalized by end-effector travel distance. Our proposed method achieves similar mechanical power efficiency to the policy trained without a perception reward, with both outperforming the FOV reward policy. The FOV reward results in inefficient base attitude and locomotion control, as it prioritizes keeping the shuttlecock within the FOV at the expense of energy efficiency.  \textbf{(D-E)} An example robot pitch trajectory during a hit. (i) ANYmal observes the shuttlecock circled in red. (ii) ANYmal pitches down while keeping the shuttlecock in the FOV. (iii) ANYmal pitches up to observe the shuttlecock for longer. (iv) ANYmal successfully hits back the shuttlecock, making it reappear in the FOV. Meanwhile, ANYmal returns to the stance posture.}
    \label{fig:active_perception}
    
    % \vspace{-1.5em}
\end{figure*}

The coordinated interplay between perception and movement was essential for the robot to successfully track and respond to the shuttlecock during gameplay. This section discusses the policy's emergent active perception behavior, evaluates its influences, and compares it to baseline methods. We evaluated the effectiveness of the proposed active perception training and compared our method with two baselines with diverse incoming shuttle trajectories in simulation with the perception noise model. The first baseline, which we call the \ac{FOV} Reward policy, was trained with an explicit reward for keeping the shuttlecock within the \ac{FOV}, with this reward term tuned to a scale similar to our proposed perception error reward for meaningful comparisons. The second baseline, the No Perception Behavior policy, observed the ground truth shuttlecock states and hence the ground truth interception position, making it not incentivized to learn perception-driven behaviors.

The perception error in this section refers to the mean $L2$-error between the true shuttlecock position and the estimated position from the \ac{EKF} at the time of the desired racket swing based on the measurement noise model. Overall, the perception error for all three methods depended on the landing location of the shuttlecock (Fig. 4A), with larger errors occurring near the edge of the badminton service area. Notably, the baseline policy trained with \ac{GT} perception demonstrated no active perception behavior, resulting in substantially larger errors than the other methods.

We measured the mean and standard deviation of perception errors by regions (a, b, c) of the badminton court (Fig. 4B). Both the proposed method and the policy trained with the \ac{FOV} reward notably outperformed the no-perception-behavior policy in region c, where the shuttlecock was more likely to exit the robot's \ac{FOV}. The similarity in perception errors between our proposed method and the-\ac{FOV} reward baseline was due to the scaling of the \ac{FOV} reward, which was adjusted to ensure that the policy still performed reasonable locomotion and racket swings. Although we could have increased the \ac{FOV} reward scaling arbitrarily to further reduce its perception error, doing so would have led to an unfair comparison, as it would have overemphasized \ac{FOV} tracking at the cost of other important behaviors. Nonetheless, the similar perception error between the two methods suggested that our approach could achieve active perception without relying on explicit \ac{FOV} rewards.

Another comparison we made was concerning normalized mechanical power, which we define as
\begin{equation}
    \varphi = \sum_{\text{all joints}} [\tau \Dot{q}]^{+}/d\,,
\end{equation}
where $\tau$ denotes the joint torque, $\Dot{q}$ is the joint velocity, and $d$ is the target distance. This metric provided an indication of the policies' energy efficiency. On this scale, our method performed comparably to the no-perception-behavior baseline, with both outperforming the \ac{FOV}-reward baseline (Fig. 4C). Important to note is that the no-perception-behavior baseline represented an upper bound of the mechanical power performance, as it solved a subset of the tasks of our proposed method in this comparison. The previously presented metrics indicated that our method balanced between active perception and efficient movement, optimizing both mechanical power and \ac{EE} tracking.

Fig. 4D and 4E illustrate an instance of the learned active perception behavior observed in our system. The robot started in a stationary position (Fig. 4D, i). Once the interception target was registered (Fig. 4D, ii), the robot first pitched down while keeping the shuttlecock in the upper part of the \ac{FOV}. Then, it pitched up (Fig. 4D, iii) to reduce the shuttle angular velocity with respect to the camera frame, thus reducing motion blur and keeping the shuttlecock in sight for longer. As soon as the shuttlecock exited the \ac{FOV}, the robot pitched down again (Fig. 4D, iv) to adjust the robot posture for the racket swing. In this instance, the active perception behavior led to \unit[0.10]{s} of additional sight of the shuttle flight. 

\subsection*{Gait adaptation}
Gait adaptation played a critical role in the robot's ability to intercept and return the shuttlecock effectively under varying distances and time constraints. This section discusses the robot's gait patterns in response to different task conditions, as illustrated in Fig. 5. The figure showcases some of the emergent adaptive behavior, with additional comparisons available in the supplementary material and the full video documentation in supplementary movie S3.

We first examined the relationship between gait and the distance the robot needed to cover to intercept the shuttlecock. For this, we conducted hardware experiments where we swept across increasing distances with a fixed time to reach the target (\unit[1.6]{s}) while keeping track of the foot contacts.

At short distances of \unit[0.5]{m}, no locomotion was necessary. The robot slightly lifted its left front (LF), right front (RF), and right hind (RH) legs to reorient the base while keeping the left hind (LH) in contact with the ground, focusing on precise positioning of the \ac{EE} for the swing. By the time of the swing, all feet were in contact with the ground (Fig. 5B).

At medium distances of \unit[1.5]{m}, the robot moved with irregular gait patterns, engaging all four legs in swing phases. The right-side legs, being farther from the target, had notably longer air time than the left. At the time of the swing, three feet remained in contact with the ground (Fig. 5C).

At longer distances of \unit[2.2]{m}, the robot employed a high-frequency gait resembling galloping between \unit[1.6]{s} and \unit[0.6]{s} before the commanded swing. As the swing time approached, the robot adjusted its gait and prepared to lift the right legs (Fig. 5A). The extended flight phase of the right legs enabled an arm extension of \unit[1.0]{m} in the direction of the target at the time of the swing. One second after swinging, the robot recovered from the dynamic pose and had all four feet in contact.

We also analyzed the gait pattern's dependency on the time the robot had to execute a maneuver. When faced with increased urgency from imminent swing targets, the robot demonstrated adaptive gaits to reach the target while prioritizing safety. The targets in this comparison were located \unit[2]{m} from the robot's initial base position in $y$-direction.

Furthermore, we observed that the emergent coordination between leg and arm motion emerged under the influence of motion regularization penalties during training. In our training framework, we applied uniform joint torque and acceleration penalty scales across all joints, resulting in the robot prioritizing base tilting and arm usage for hitting nearby targets. By reducing regularization weights on the legs, we could encourage more dynamic leg movements, as demonstrated in the supplementary material fig.~S7.

When given \unit[0.8]{s} to reach the target, the robot stepped with high frequency with the LF, RF, and LH feet while only making one long step with the RH leg. By extending the arm it managed to successfully reach the swing target in time. Under harder time constraints of \unit[0.4]{s}, it was physically impossible for the robot to reach the target. Despite its attempt, the robot failed to reach the required position, resulting in a missed hit. However, it managed to avoid excessive base limb motion, showcasing the policy's robustness even when faced with unreachable commands and demonstrating awareness of its current physical limitations.

\begin{figure*}[htp!]
    \centering
    \includegraphics[height=0.77\textheight]{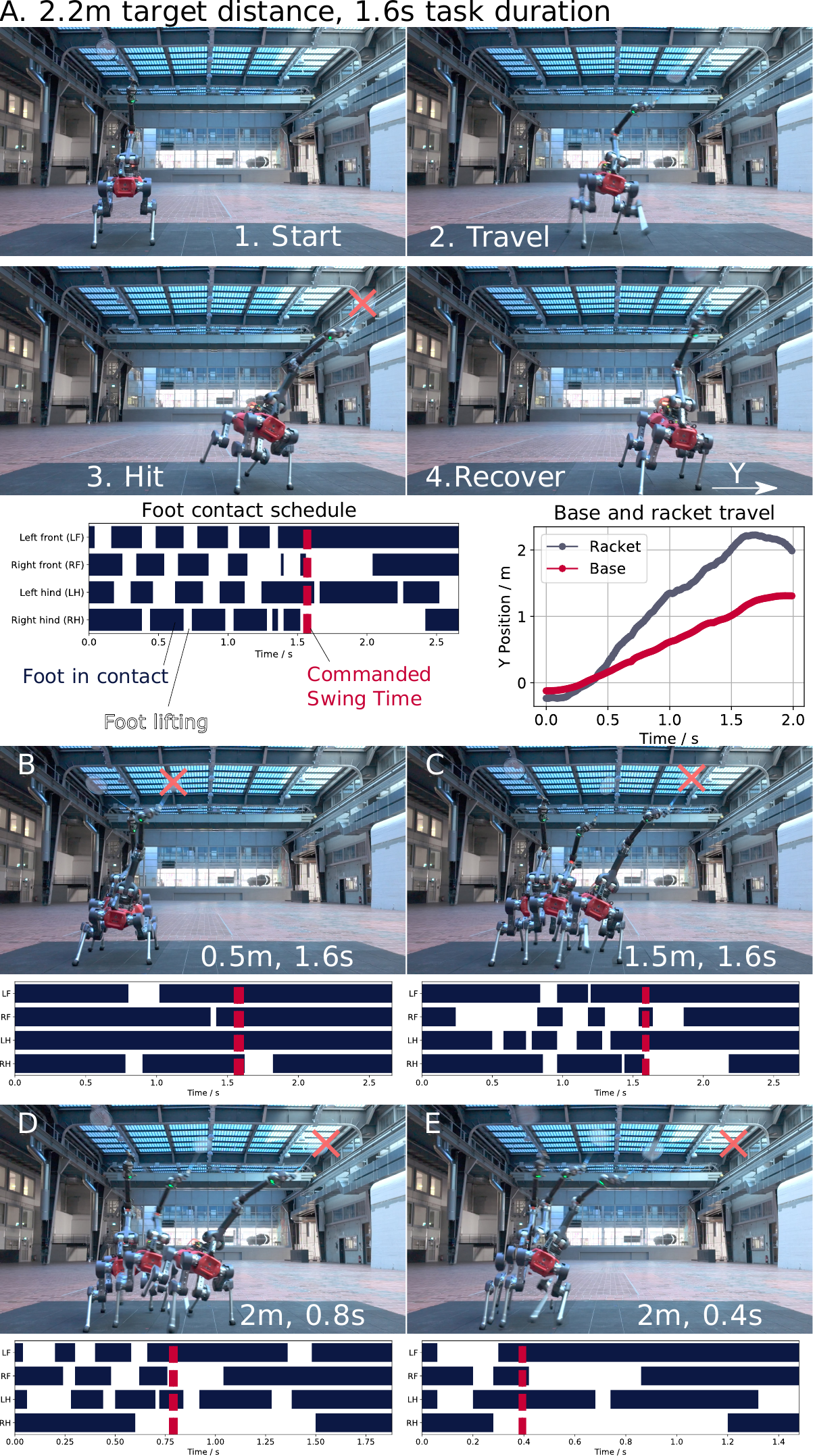}
    \caption{\textbf{Gait adaptation based on distance and time urgency. (A)} The robot reaches targets \unit[2.2]{m} away from the starting base position in \unit[1.6]{s} using a galloping-like gait. Near the swing, the longer right foot lift phase and increased $y$-coordinate difference between the base and racket indicate a gait adjustment. \textbf{(B-C)} When targeting nearby positions, the robot barely lifts its feet. It steps to locomote when the target is out of reach. \textbf{(D-E)} Under tight time constraints, the control policy balances between maintaining safety and tracking the target accurately.}
    \label{fig:gait_adaptation}
    
    % \vspace{-1.5em}
\end{figure*}

% what's y axis
% initial position, hit position

\section*{Discussion}
We present a legged manipulator system capable of playing badminton using only onboard perception. This system showcases advancements in coordinating legged locomotion with manipulation and balancing limb agility with perception accuracy, highlighting its potential in dynamic and competitive human sports. Through the use of multi-target training and asymmetric actor-critic \ac{RL}, coupled with a perception model, the robot was able to develop sophisticated human-like badminton behaviors. These include follow-through after hitting the shuttlecock and active perception to enhance shuttle state estimation, achieved without explicit training heuristics.

The robot's performance was extensively evaluated through various hardware experiments, including success rate assessments, tasks involving targets at different distances and under varying time constraints, and verification of active perception behavior. During multiple collaborative games with humans in different environments, the system demonstrated its ability to respond to shuttle shots with varying angles, speeds, and landing locations, achieving ten consecutive shots in a single rally under mildly windy outdoor conditions.

Incorporating the noisy perception model and using the same \ac{EKF} for both training and deployment establishes a consistent mapping between motion history and expected perception outcomes across simulation and hardware. This provides a means to address a known limitation of privileged learning (teacher-student training): the information gap between the teacher policy trained with perfect perception and the student policy for deployment. In such a framework, the teacher policy has no incentive to learn active perception behaviors because it already has access to perfect observations. The student policy -- trained through behavior cloning -- only mimics these actions based on partial observations and a latent vector reconstructed from proprioception and perception histories. As a result, neither policy develops active perception behaviors, and a discrepancy arises in the information used for control between the two policies. Our method bridges this gap by encouraging active behaviors through the aforementioned motion-perception mapping of incorporating the \ac{EKF} also in training. This approach could be further extended by replacing the regressed model with learned perception models to enhance generalizability. Although this may introduce additional training complexity and computational overhead, it presents an exciting direction for future research in improving active perception learning within reinforcement learning frameworks.

We identify several other promising extensions to further enhance the robot's athletic capabilities. Currently, a set of configurable rules determines the swing height, velocity, and orientation. Although the robot's high degree of freedom offers substantial potential for more nuanced racket control, these configurable rules underutilize this capacity. A high-level badminton command policy that adapts swing commands based on the opponent's body movements could improve the robot's ability to maintain rallies and increase its chances of winning. Furthermore, the current control policy is trained to hit interception targets between \unit[0.9-1.4]{m} above the robot's base using the same side of the racket. Diversifying the swing motion by extending the training scheme could further enhance performance. Moreover, although the policy performs well across shot directions, success rates are lower when returning shuttlecocks that land behind the robot. This limitation stems primarily from perception constraints, as giving the robot ground-truth perception, as shown in Fig. 3A, makes the performance almost symmetric. Having to maintain the shuttlecock within the \ac{FOV} becomes notably harder when walking backward. A wider \ac{FOV} camera or an actuated camera pitch joint could mitigate this issue.

Additionally, the current system relies heavily on an \ac{EKF} applied to a single off-the-shelf stereo camera for shuttlecock state estimation. This approach could be refined by integrating additional sensing modalities, such as torque and sound for impact detection, or incorporating extra RGB, depth, or event-based cameras to enhance the robot's response to physical interactions during more intense gameplay—such as when trying to hit back smash shots. Since human players often predict shuttlecock trajectories by observing their opponents' movements, human pose estimation could also be a valuable modality for improving policy performance.

In conclusion, our research demonstrates that a legged mobile manipulator can autonomously play with human players in a full-scale sport by tightly coupling whole-body maneuver with perception inside a single \ac{RL} framework. Embedding the parameterized perception model and the same \ac{EKF} used on hardware in training allows the robot learns to reduce observation error while executing agile strikes. Further simulated experiments hint potential extension of the proposed framework to other legged manipulator morphologies, such as humanoids (supplementary movie S5). Beyond badminton, the method offers a template for deploying legged manipulators in other dynamic tasks where accurate sensing and rapid, whole-body responses are both critical.

\section*{Materials and Methods}

\begin{figure*}[ht!]
    \centering
    \includegraphics[width=0.8\textwidth]{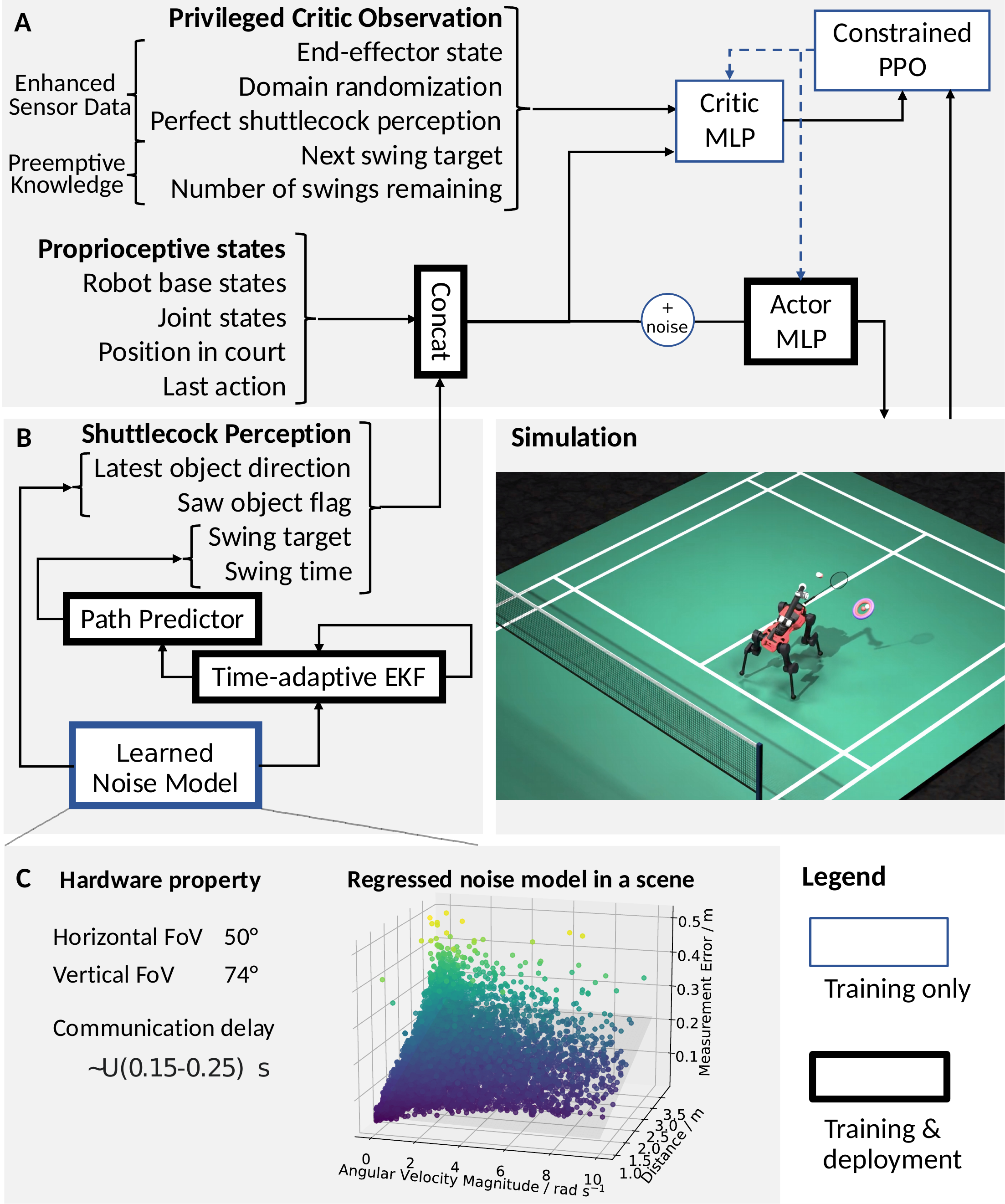}
    \caption{\textbf{Overview of the training method. (A)} Joint control policy training with RL. The policy is trained with an asymmetric actor-critic, with privileged environment states and MDP information given only to the critic network. The policy receives the noised proprioceptive states and shuttlecock perception with simulated noise. \textbf{(B)} The shuttlecock perception module in simulation. The time-adaptive EKF and the path predictor are reused during the deployment. During the training, we used a regressed perception noise model based on real camera noise collected from the hardware. \textbf{(C)} Object perception noise model. The object is in the simulation with the regressed detection probability and measurement noise if it is in the camera FOV.}
    \label{fig:training_pipeline}
\end{figure*}

The primary goal of our system was to perceive the shuttlecock, compute the swing target, and execute the swing motion. An overview of our method is presented in Fig.~6.

\subsection*{RL-based dynamic whole-body visuomotor skills}
We trained the robot's whole-body maneuvering policy using \ac{RL} in a high-fidelity simulated environment. This environment included detailed robot dynamics, such as manipulator transmission modeling, joint actuator modeling, and dynamics parameters obtained through system identification. Additionally, constrained \ac{RL} was used to enforce hardware constraints specific to the robot. To further improve transferability to the physical robot, we applied domain randomization techniques, such as varying friction coefficients, adding base masses, and introducing occasional random pushes. More detailed explanations of the training environment implementation are provided later in this section and in the supplementary material.

\begin{figure*}[htp!]
    \centering
    \includegraphics[width=0.9\textwidth]{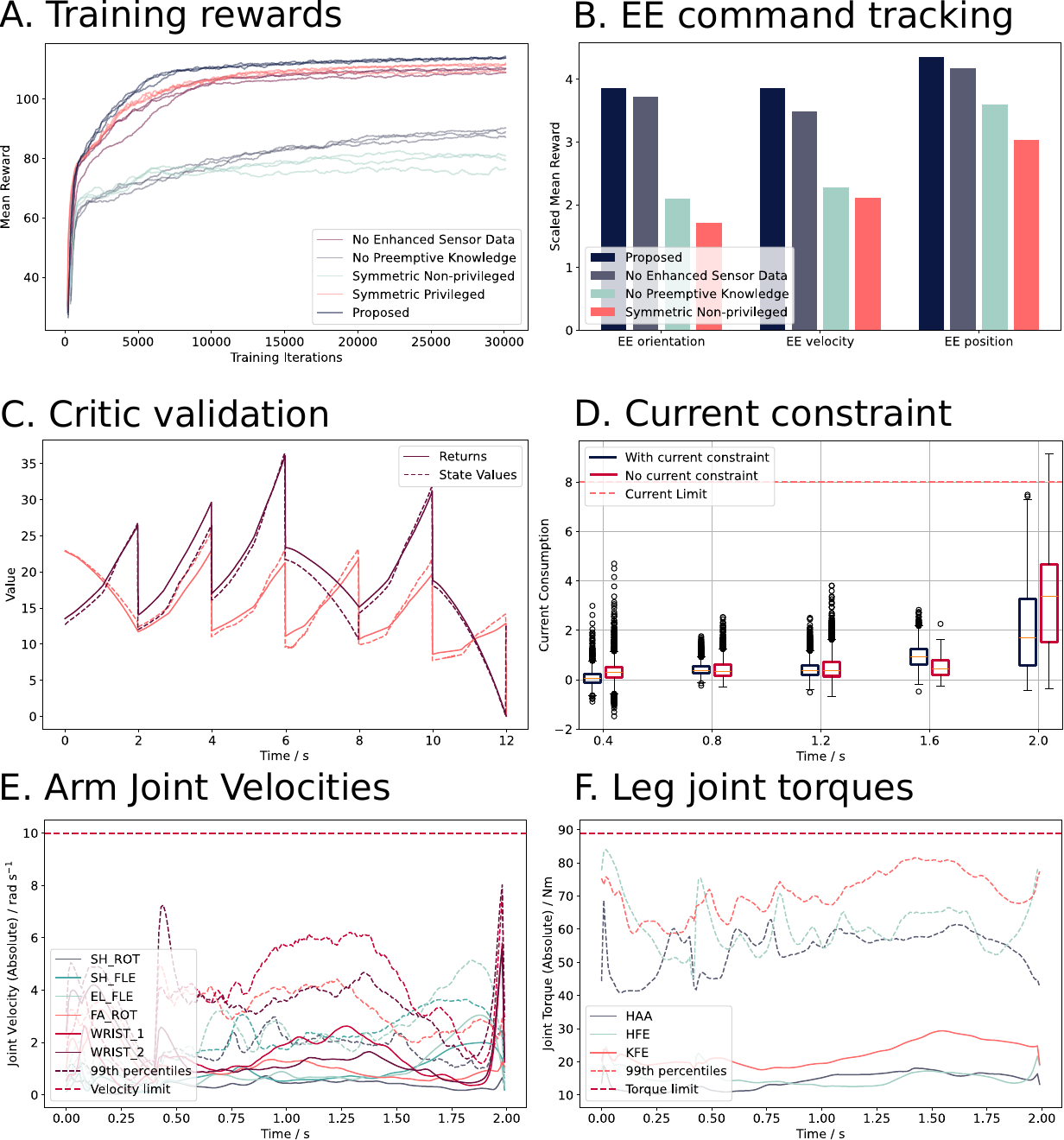}
    \caption{\textbf{Validation of the training method. (A-B)} Ablation studies comparing different observation configurations show that our method consistently outperforms baseline training setups in terms of both convergence time and final end-effector tracking performance. \textbf{(C)} Multiple sample trajectories (represented by different colors) demonstrate that our value prediction closely aligns with the computed trajectory return. \textbf{(D)} The current consumption constraint is respected by policies trained using the N-P3O formulation. \textbf{(E)} The arm joint velocities over the sample swings. \textbf{(F)} The leg joint torques over the sample swings. \textbf{(D-F)} The box and line plots use data from all target positions on the court, each totalling 199,650 trajectories.}
    \label{fig:method_validation}
    
    % \vspace{-1.5em}
\end{figure*}

% add badminton shuttlecock trajectory

To achieve \ac{EE} swing tracking and allow the policy to learn post-swing follow-through behaviors, we simulated six swing targets per episode. However, this implementation made the state value function dependent on the number of hits remaining, though this information should not be made available to the deployed policy actor. To address this, we employed an asymmetric actor-critic approach~\cite{pinto2017asymmetricactorcriticimagebased} with time-based rewards~\cite{rudin2022advanced} for training, as shown in Fig. 6A. In this setup, the critic network was provided with additional information, such as the number of remaining hits in the episode, to better learn the value function. The actor network only received the robot states and swing target data, which included simulated noise. Table S2 in the supplementary materials presents a detailed list of observations.

We categorized the additional critic observations into two types: enhanced sensor data and preemptive knowledge. Enhanced sensor data included higher-quality versions of existing observations, such as noiseless base and joint states, as well as the \ac{EE} states. However, the \ac{EE} states were also derivable via forward kinematics from joint states, albeit with noise. Preemptive knowledge referred to information used to define the MDP that was unavailable during deployment as it depended on the opponent's action, such as the number of remaining targets and the distance between the current target and the next target.

Both categories of observations enhanced the critic network's accuracy in estimating the state value function. Enhanced sensor data reduced stochasticity in the \ac{N-P3O} policy gradient, and preemptive knowledge completed the set of state variables upon which the value function relied. As detailed in the following section, the training environment was structured around consecutive target swings, with rewards primarily based on \ac{EE} state tracking. The number of remaining swings in an episode substantially influenced the expected return from the current state. Additionally, the distance between consecutive targets provided the critic with insight into the anticipated motion vigor and tracking precision required, supplying predictive information that further refined return estimation and, consequently, improved policy training.

Individual ablation studies (Fig. 7A and Fig. 7B) show that preemptive knowledge contributed to both the learning process and the final policy performance in \ac{EE} tracking. The proposed observation format outperformed symmetric privileged training (teacher training in privileged learning~\cite{lee2020learning}). This improvement was due to the actor policy in symmetric training developing unnecessary time-dependent behaviors, which introduced additional challenges for value function estimation.

The critic's value function predictions closely match the trajectory return computed from the discounted rewards cumulatively summed from the end, validating the effectiveness of the enhanced critic training (Fig. 7C). In contrast, removing the additional critic observations resulted in larger value function errors, leading to worse learning outcomes, as shown in Fig. 7B.

During training, we modeled the perception detection probability and noise as functions of the distance to the shuttlecock, the robot's angular velocity, and whether the shuttlecock was within the robot's FOV. This perception model was regressed from data collected using the robot hardware. We then utilized the same \ac{EKF} and shuttlecock trajectory prediction module both during training and deployment, ensuring consistency, as depicted in Fig. 6B.

\subsection*{Swing tracking}
We employed a time-based swing reward mechanism to incentivize accurate and timely racket swings. This included rewards for position, orientation, and velocity, activated for a single timestep per swing. The orientation reward specifically targeted the angle difference from the normal direction to the racket face.

During training, the robot observed the swing target position as the relative position between the \ac{EE} and swing target, expressed in the base frame of the robot. This observation formulation helped maintain robust tracking when domain randomization was applied to the arm dynamics, which suggested smaller sim-to-real transfer challenges. As we expected flat terrain during the deployment, we did not implement advanced terrain types or curriculum during the training. An overall flat terrain with small unevenness was used to encourage higher foot-lifting, which helped the sim-to-real transfer. The supplementary materials provide additional training details, including other observations, network architecture, and hyperparameters.

\subsection*{Perception noise model}
To fit the perception noise model, we collected data where the camera moved around while observing a fixed shuttlecock at a known location. The camera's position was tracked using a motion capture system, allowing us to compute the distance and angular velocity between the camera and the shuttlecock. Supplementary movie S4 shows the data collection procedure. The detection probability and the shuttlecock position error were regressed as a linear function of shuttle distance and angular velocity, providing a noise model that could be deployed during the training with negligible computation overhead. An example of the regression is depicted in Fig. 6C. 

Both our shuttlecock trajectory generation and the \ac{EKF} followed the shuttlecock dynamics model \cite{cohen2015physics} with a measured aerodynamic length $L$ of \unit[4.1]{m}. This aerodynamic length is defined as $L=2m/\rho SC_D$, where $m$ is the projectile mass, $\rho$ is the air density, $S$ is the cross-sectional area, and $C_D$ is the air drag coefficient.
\begin{equation}
    m\frac{d\mathbf{v}}{dt} = m\mathbf{g} - m||\mathbf{v}||\frac{\mathbf{v}}{L}
\end{equation}
where $m$ is the mass of the shuttlecock, $\mathbf{v}$ is its velocity, and $\mathbf{g}$ is gravitational acceleration. The process noise and measurement noise configurations are available in the supplementary material.

During training, we integrated the perception noise model, the \ac{EKF}, and shuttlecock trajectories into the learning loop. We first generated shuttlecock trajectories with random initial states and aerodynamic lengths. For each target swing in the training environment, we sampled a target swing height and a shuttle trajectory from the saved trajectory pool. We padded and translated the trajectory so the shuttle reached the target swing height at the commanded swing time. The shuttle detection and measurement error was sampled using the regressed noise model (Fig.6C) and given to the \ac{EKF}. Because the same EKF and trajectory prediction module are employed during training and on hardware, the single-frame noise introduced here is filtered identically in both cases, capturing not just per-step measurement errors but also the final interception prediction error. This enabled direct penalization of observation error, rather than imposing hard-coded \ac{FOV} constraints, allowing the RL algorithm to naturally balance perception accuracy and motion control for active perception learning.

To avoid the computation cost required to rollout the full shuttle prediction, we computed the final target offset linearized with respect to the current state estimation error based on the \ac{EKF} estimation and the noiseless shuttlecock trajectory. Details on the trajectory distribution and the target offset approximation are presented in the supplementary material.

This approach modeled the perception noise and reused the filter and prediction modules deployed on the hardware to reflect real-world conditions. Note that the perception noise level is subject to the testing site's light conditions and ambient color. We acknowledged that this is a limitation of our approach and that there is an expected perception error difference when deploying the trained policy in a new environment. However, the learned active perception behavior shown in Fig. 4. would still qualitatively transfer to different environments and decrease the perception error.

\subsection*{Training}
The training process used the IsaacGym simulator~\cite{makoviychuk2021isaac} with the legged\_gym framework~\cite{makoviychuk2021isaac, rudin2022learning}. The training used the \ac{N-P3O}~\cite{lee2023evaluation}, a constrained variant of \ac{PPO} algorithm~\cite{schulman2017proximalpolicyoptimizationalgorithms}. The policy approached the maximum training reward after around 7500 iterations (Fig. 7A), corresponding to \unit[4.81]{h} wall-clock time on a single RTX 2080Ti GPU. For deployment, the policy was usually trained for $1$ - $2$ days for better convergence.

\subsection*{Perception deployment}
In the deployment phase, we implemented several key components to ensure the accurate tracking and striking of the shuttlecock. We utilized a color-based filtering approach for effective shuttlecock tracking in the camera frame. Specifically, we employed the \ac{HSV} scale to filter out the shuttlecock's orange color by setting an upper and lower range for the \ac{HSV} values. This enabled us to isolate the shuttlecock from the background effectively. Using the stereo information provided by the ZED X camera, we then transformed the filtered positions from 2D image coordinates to the robot's map frame.

The map frame -- a globally consistent world frame generated by the robot's \ac{SLAM} pipeline, distinct from the odometry frame, which can accumulate drift over time -- essential for accurate localization and tracking, was derived through the integration of modular sensor fusion (MSF)~\cite{lynen13robust} and CompSLAM~\cite{khattak2020complementary}. A stable and accurate map frame that properly accounted for the robot's movements was critical, as the shuttlecock's state estimation was filtered within this frame. Any drift in the map frame would have resulted in a noisy shuttlecock estimation or directly led to incorrect interception positions when transformed into the base frame, thereby affecting the swing command observations.

Due to notable base angular velocity during the swing preparation phase, accurate timing information on the shuttle's position was required to determine its position in the world frame. For this purpose, the ZED X camera firmware provided synchronized image timestamps. Camera selection also played a key role in our perception system. We opted for a narrower \ac{FOV} instead of a wide-angle camera to enhance angular resolution. This choice reduced measurement noise and improved the accuracy of shuttlecock tracking, particularly during high-speed motions where precise angular data is critical. For our perception system, we measured a total of \unit[60-160]{ms} delay between the camera shutter time and when the shuttle's positions could be computed.

Once the shuttlecock's position in the map frame was obtained, it was processed by an \ac{EKF} with parameters identical to those used during training. The \ac{EKF} output a filtered shuttlecock state estimate, enabling trajectory prediction and interception point computation. In both simulation and real-world deployment, if the shuttlecock exited the robot's \ac{FOV}, the system maintained the last predicted interception position for up to \unit[2]{s}, during which the robot attempted to strike based on this estimate.

\subsection*{Sim-to-real practicalities}
Several practical considerations were addressed during hardware deployment. Unlike the actuator network model used for the legs~\cite{hwangbo2019learning}, we applied a Covariance Matrix Adaptation Evolution Strategy (CMA-ES) to optimize the hardware model parameters~\cite{filip2024} in IsaacGym. This approach was chosen because the manipulator we used did not provide accurate torque measurements, and it used \acp{QDD}, which have more transparent dynamics compared to the \acp{SEA} on the legs. We collected robot joint position trajectories resulting from sine wave command trajectories of varying frequency and swing command trajectories on the hardware. We optimized for the model parameters, including the joint friction, damping, and armature, to match the hardware joint position trajectory under identical commands.

The arm current consumption was limited to \unit[8]{A} on the ANYmal robot by a fuse. Constrained \ac{RL} technique N-P3O \cite{lee2023evaluation} were used to enforce arm current consumption constraints on the robot.
\begin{equation}
    |I_{\text{total}}| < \unit[8]{A}\,,
\end{equation}
where the total current $I_{\text{total}}$ is calculated by summing the contributions from resistive power, derived from the motor constant $K_m$ and the mechanical power, computed from the torque $\tau_{i}$ and motor velocity $\omega_{i}$, and dividing by the voltage $V$
\begin{equation}
    I_{\text{total}} = \sum_{i=1}^{N} \left( \frac{\tau_{i}}{K_{m}} \right)^2 \frac{1}{V} + \sum_{i=1}^{N} \frac{\tau_{i} \cdot \omega_{i}}{V}\,.
\end{equation}
The policy avoided the current limit with the N-P3O constraint implementation, in contrast, the baseline policy without the constraint violated the constraint even with soft over-current penalties. During the hardware deployment, the policy trained with N-P3O never violated the constraint.

The actuator torque and velocity constraints were included as soft penalties in the reward function, as these were also treated as soft constraints on the hardware. Although exceeding these limits wouldn't cause immediate failures, frequent violations could lead to wear or damage. The distributions of maximum arm drive velocity and leg drive torque in our test scenario are shown in Fig. 7E and Fig. 7F, respectively. Fig. 7E indicates fast motion is observed in the wrist drives both in the backswing phase and during the swing, whereas the leg torque usage remained more consistent throughout the phases. During all phases, these values approached their limits but remained within them.

\subsection*{Statistical analysis}
Statistical analyses were performed in Python using the NumPy library to compute means and standard deviations. Data were sampled at 100 Hz for simulated experiments and 400 Hz for hardware experiments. The analyses shown in Figs. 3A-B, 4A-C, and 7D-F used 1650 trajectories per target position, with perturbed initial robot joint configurations. For the normalized mechanical power computation (Fig. 4C), trajectories with zero target distance were excluded to prevent division by zero. The mean training rewards shown in Fig. 7A-B were averaged across 4096 parallel training environments using three random seeds. All plots in Fig. 7 were generated from simulated data using Matplotlib, with a convolution filter of window size 50 applied to Fig. 7A for smoothing.

\section*{Acknowledgments}
\textbf{Acknowledgements:} We thank Changan Chen and Kaixian Qu for dedicating extensive time as the robot's opponents. We also thank Nikita Rudin for initial project discussions and Fabian Tischhouser for extensive hardware engineering support. We are grateful to Dylan Vogal for his insightful feedback on Movie 1. Additional thanks to Junzhe He, Tianxu An, Jan Preisig, Fan Yang, Mayank Mittal, Yanqing Shen, Takahiro Miki, Filip Bjelonic, and Jia-Ruei Chiu for their assistance in experiments and data collection, and to Eris Sako for hardware insights. We acknowledge Kento Kawaharazuka, David Hoeller, and Joonho Lee for project discussions, and Emre Elbir, Ennio Schnieder, Andreas Binkert, Andri Graf, Johann Schwabe, Flurin Schindele, and Laurin Schmid for CAD and infrastructure support. We also used ChatGPT and DeepSeek to assist with revising and refining the language in this paper. \textbf{Funding:} This work was supported by Intel Labs, the Max Planck ETH Center for Learning Systems, and the National Centre of Competence in Research Robotics (NCCR Robotics). Additionally, this work was conducted as part of ANYmal Research, a community dedicated to advancing legged robotics. \textbf{Author contributions:} Y.M.: Conceptualization, simulation, sensor selection, data collection, policy training, experiments, investigation, analysis, visualization, writing. A.C.: Sensor selection, substantial revision. F.F.: Initial project discussions, revision. M.H.: Initial task scope, sensor selection, resources, supervision, revision. \textbf{Competing interests:} There are no competing interests to declare. \textbf{Data and materials availability:} All (other) data needed to evaluate the conclusions in the paper are present in the paper or the Supplementary Materials. The data for this study have been deposited in the database DOI: \href{https://doi.org/10.5281/zenodo.15242151}{10.5281/zenodo.15242151}.

\newpage
\newpage

%%%%%%%%%%%%%%%% START OF SUPPLEMENT %%%%%%%%%%%%%%%

% Figures, tables, equations and pages in the supplement are numbered S1, S2 etc.
\renewcommand{\thefigure}{S\arabic{figure}}
\renewcommand{\thetable}{S\arabic{table}}
\renewcommand{\theequation}{S\arabic{equation}}
\renewcommand{\thepage}{S\arabic{page}}
\setcounter{figure}{0}
\setcounter{table}{0}
\setcounter{equation}{0}
\setcounter{page}{1} % not 0 as \newpage already started a supplementary page
% References continue the numbering from the main text.

%%%%%%%%%%%%%%%% SUPPLEMENT TITLE PAGE %%%%%%%%%%%%%%%

\section*{Supplementary Materials}

% Fill out the numbers for each type of supplementary material,
% and delete any lines that aren't applicable.
% These are just example numbers that don't match the rest of this template.

%%%%%%%%%%%%%%%% MATERIALS AND METHODS %%%%%%%%%%%%%%%

\subsection*{Nomenclature}

\begin{table}[h!]
\centering
\begin{tabular}{cl}
\rowcolor[HTML]{EFEFEF} 
\textbf{Symbol} & \textbf{Discription}           \\ \hline
$q$, $\Dot{q}$               & Joint position, joint velocity \\
$\hat{q}$, $\Dot{\hat{q}}$               & Motor position, motor velocity \\
$I$               & Current \\
$V$               &  Voltage \\
$\tau$               &  Torque \\
$K_m$               &  Motor constant \\
$\mathbf{p}$               & Body position                  \\
$\mathbf{v}$               & Body velocity                  \\
$\mathbf{a}$               & Body acceleration              \\
$\omega$               & Angular velocity              \\
$r$               & reward                         \\
$\mathbf{n}$                & Normal vector                  \\
$T$              & Racket swinging time           \\
$t_k$               & Current time                   \\
$\sigma$                & Reward sensitivity factor      \\
$S_C$                & Cosine similarity                \\
$L$                & Aerodynamic length          \\
$\epsilon$                & Perception error          \\
$\varphi$                & Normalized mechanical power          \\
$\mathcal{D}$                & Detecting the shuttlecock, binary          \\
$\mathcal{T}$                & The environment is given a swing target, binary  
\end{tabular}
\end{table}

\subsection*{Deployment Interception criteria}

\begin{figure}
    \centering
    \includegraphics[width=0.9\linewidth]{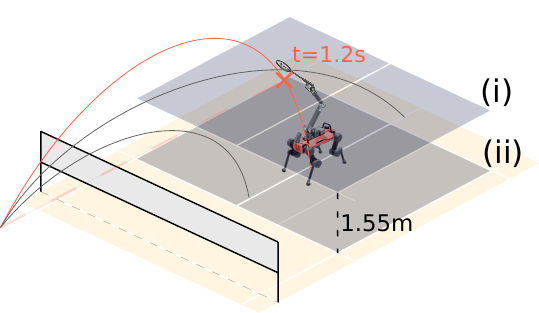}
    \caption{ \textbf{Shuttlecock trajectory qualification heuristics for hardware deployment.}  The robot only hits back shuttlecocks with flight trajectories crossing the badminton service area both at ground height and 1.55m above the ground (e.g. the orange trajectory). The swing time is computed based on the time that the trajectory crosses a configurable height.}
    \label{fig:deployment_heuristics}
\end{figure}

During hardware experiments, we qualified the predicted shuttlecock trajectory to intercept if it intersected both rectangles (i) and (ii) in fig. S1. In this context, only the orange trajectory qualified, as it crossed both rectangles, whereas the other two trajectories only intersected one rectangle each. Rectangle (i) was positioned \unit[1.55]{m} above the ground, corresponding to the height of a standard badminton net, and rectangle (ii) was at ground level. Extending the rectangles to cover the entire single-player court would have aligned the scoring qualifications with formal human matches. However, we limited the region to the service area as we expected the robot to have a higher probability of successfully returning the shuttlecock in this area (Fig. 3).

\subsection*{Belt Transmission Modeling}

\begin{figure}
    \centering
    \includegraphics[width=0.8\linewidth]{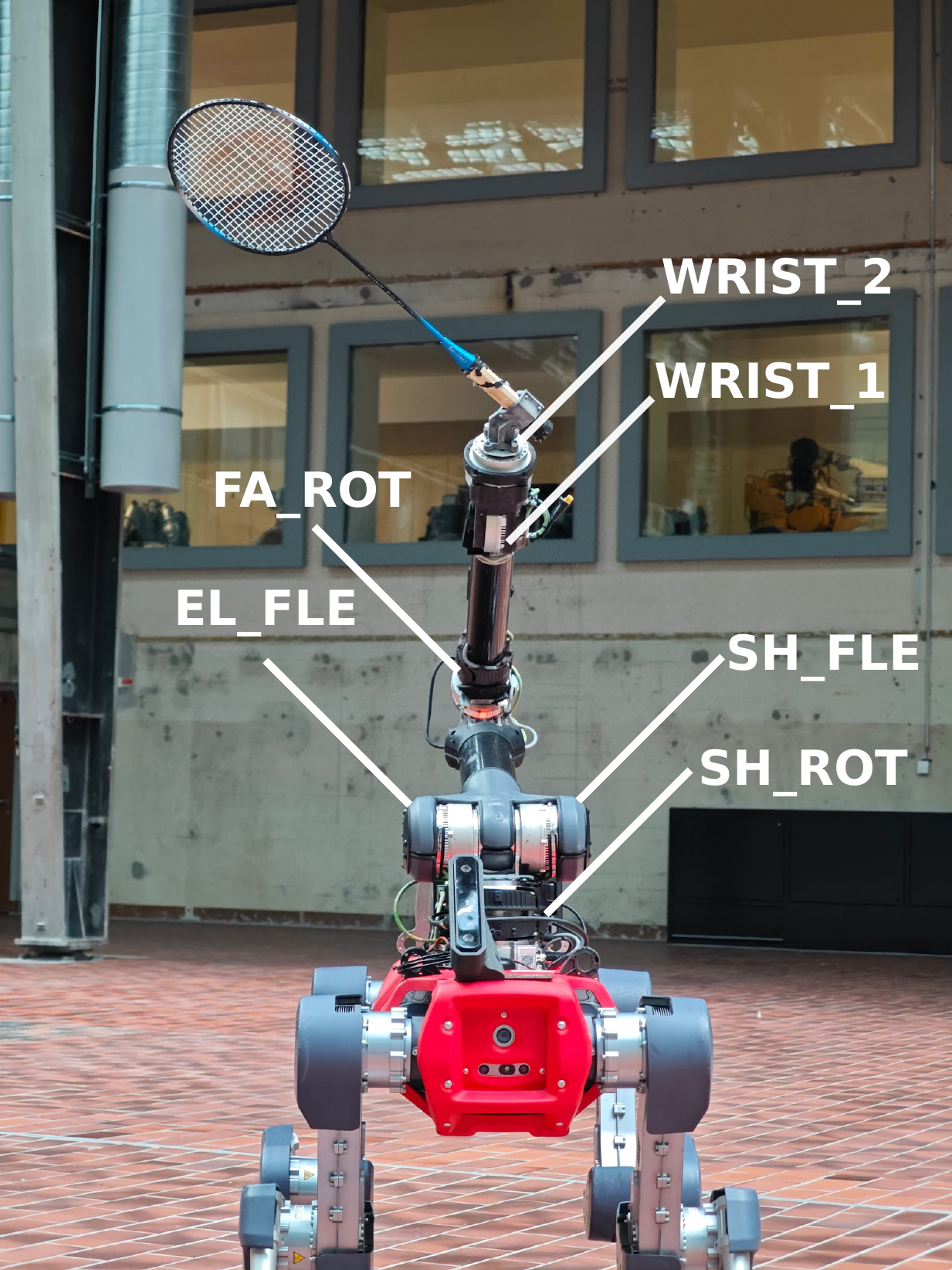}
    \caption{Motor names of the DynaArm manipulator.}
    \label{fig:arm_drives}
\end{figure}

The manipulator used in this project, DynaArm (shown in fig.~\ref{fig:arm_drives}), features a belt transmission system that controls the elbow flexion joint via a motor located at the shoulder with a gear ratio of 1:1. This transmission setup is not natively supported in the IsaacGym simulator and introduces major changes to the robot's dynamics. To address this, we implemented a serial-DynaArm conversion to account for these dynamic differences. Here, $\delta \Tilde{q}$ and $\delta \Tilde{\Dot{q}}$ represent the motor errors, which are distinct from the joint errors $\delta q$ and $\delta \Dot{q}$. In our case, the desired joint velocities $\Dot{q}^{des}$ were set to zero for all joints.

\begin{equation}
    \delta \Tilde{q}_{EL\_FLE} = (q^{des}_{EL\_FLE}-q_{EL\_FLE}) + (q^{des}_{SH\_FLE}-q_{SH\_FLE})
\end{equation}

\begin{equation}
    \delta {\Dot\Tilde{q}}_{EL\_FLE} = (\Dot{q}^{des}_{EL\_FLE}-\Dot{q}_{EL\_FLE}) + (\Dot{q}^{des}_{SH\_FLE}-\Dot{q}_{SH\_FLE})
\end{equation}

Once the torques were computed from the correct motor errors, we added the motor torque from EL\_FLE motor to SH\_FLE to retrieve the effective joint torque in serial configuration.

\subsection*{Perception Parameters}

The parameters for the camera, shuttlecock \ac{HSV} filter, and the \ac{EKF} are listed in table~\ref{tab:perception_params}. The \ac{HSV} values are specified according to the OpenCV convention \cite{opencv_library}.

\begin{table}
    \centering
    \caption{Perception Parameters}
    \begin{tabular}{ccc}
    
    \rowcolor[HTML]{EFEFEF} 
                                                                                                     & \textbf{Parameter}   & \textbf{Value}                                                                                                                                        \\ \hline
                                                                                                     & Camera fps           & 60                                                                                                                                                    \\
                                                                                                     & Exposure             & \begin{tabular}[c]{@{}l@{}}1e-2 s (indoor, only one floodlight) \\ 4.17e-3 s (indoor, good light condition)\\ 8.33e-5 s (outdoor, sunny)\end{tabular} \\
    \multirow{-3}{*}{Camera}                                                                         & Gain                 & 100                                                                                                                                                   \\
    \rowcolor[HTML]{EFEFEF} 
    \cellcolor[HTML]{EFEFEF}                                                                         & Hue                  & \textless 5 or \textgreater 176                                                                                                                       \\
    \rowcolor[HTML]{EFEFEF} 
    \cellcolor[HTML]{EFEFEF}                                                                         & Saturation           & \textgreater 60                                                                                                                                       \\
    \rowcolor[HTML]{EFEFEF} 
    \multirow{-3}{*}{\cellcolor[HTML]{EFEFEF}\begin{tabular}[c]{@{}l@{}}Color\\ Filter\end{tabular}} & Value                & \textgreater 160                                                                                                                                      \\
                                                                                                     & Process noise std    & 2e-3                                                                                                                                                  \\
                                                                                                     & Measurement noise    & 4e-2                                                                                                                                                  \\
    \multirow{-3}{*}{\ac{EKF}}                                                                            & Reset time threshold & 0.2 s                                                                                                                                                
    \end{tabular}
    \label{tab:perception_params}
    \end{table}

\subsection*{Shuttle Trajectory Sampling}

\begin{figure}
    \centering
    \includegraphics[width=0.8\linewidth]{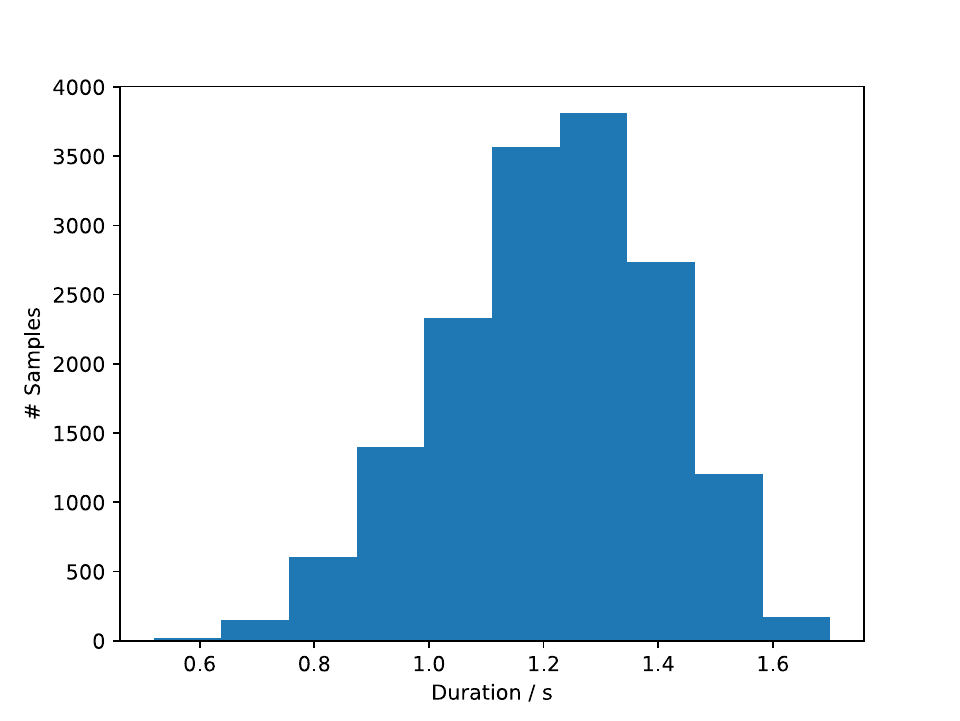}
    \caption{Distribution of the precomputed shuttlecock trajectory duration.}
    \label{fig:precomputed_duration}
\end{figure}

For policy training, shuttlecock trajectories were pre-sampled randomly, allowing us to efficiently determine the interception location and shuttle position at each timestep during training episodes with minimal computational cost. These trajectories were sampled from the following distribution (in SI units):

\begin{align*}
    p_{x, t_0} & \sim U(6, 7) \\
    p_{y, t_0} & \sim U(-2, 2) \\
    p_{z, t_0} & \sim U(-0.5, 2.5)  \\
    v_{x, t_0} & \sim U(-19, -13)  \\
    v_{y, t_0} & \sim U(-3, 3)  \\
    v_{z, t_0} & \sim U(9, 15)
\end{align*}

This distribution was designed so that shuttlecock trajectories originated near the center of the opponent's court and crossed to the robot's side at an average position of $(p_{x,T}, p_{y,T}) \approx (0, 0)$ with a height of 1.8 meters. The landing positions were spread across the entire court. The distribution of trajectory durations, from shuttle launch to interception, is shown in fig.~\ref{fig:precomputed_duration}.

For the evaluations in the Results section, we used the mean values of the distribution to simulate a nominal shuttlecock flight, adjusting the starting position to assess interception performance across different court locations.

\subsection*{Training Rewards}

We trained the badminton policy in simulation using the rewards and corresponding scaling factors listed below. Selected reward formulations are provided in Eq.~\ref{eq:rew_ee_pos} to \ref{eq:rew_perception}. Other reward terms are based on their corresponding $L2$-norms. The reward terms and scales are shown in table~\ref{tab:rewards}.

\begin{table}
    \centering
    \caption{Training Rewards}
    \begin{tabular}{cc}
    \cellcolor{grey}\textbf{Reward term}              & \cellcolor{grey}\textbf{Scale} \\
    \hline
    EE position tracking     & 6400  \\
    EE orientation tracking  & 1200  \\
    EE swing velocity        & 1200  \\
    perception error         & 3     \\
    face the net             & 0.5   \\
    torques                  & -1e-5 \\
    joint acceleration       & -1e-6 \\
    action rate              & -0.03 \\
    collision                & -2    \\
    joint position limit     & -1    \\
    joint torque limit       & -1e-3 \\
    stand still if no target & 16   
    \end{tabular}
    \label{tab:rewards}
    \end{table}

\subsubsection*{Task Rewards}

The \ac{EE} position tracking reward encouraged the racket's sweet spot to match the target interception point at the time of the swing. It was activated only for a single step per shuttle interception, at $t_k = T$.

\begin{equation}
    r_{p} = \delta (t_k-T) \frac{1}{1+||p_{EE}^*-p_{EE}||_2/\sigma_{p}} \label{eq:rew_ee_pos}
\end{equation}

Similarly, the \ac{EE} orientation tracking reward and the swing velocity reward encouraged accurate orientation and velocity tracking at the same interception timestep. The orientation reward penalized the squared cosine distance between the commanded racket-facing direction and the executed direction. The swing velocity reward was computed based on the squared difference between the commanded and actual racket sweet-spot velocity.

\begin{equation}
    r_{q} = \delta (t_k-T) \frac{1}{1+S_C({n^*_{EE}-n_{EE}})^2}
\end{equation}

\begin{equation}
    r_{v} = \delta (t_k-T) \frac{1}{1+(v_{EE}^*-v_{EE})^2/\sigma_{v}}
\end{equation}

The perception error reward was a temporally dense reward function, calculated based on the ground-truth interception position and the interception position estimated by the \ac{EKF} shuttlecock state.

\begin{equation}
    r_{\epsilon} = \frac{1}{1+||p_{i} - p^*_{i}||_2} \label{eq:rew_perception}
\end{equation}

\subsubsection*{Regularization Rewards}

The impact reward was designed to reduce stomping behavior when the robot traversed the badminton court. It lowered the reward if excessive acceleration occurred in the robot links along the $z$-axis.

\begin{equation}
    r_{impact} = \sum^{links} ||a_{z,link}||^2
\end{equation}

Additionally, in environments where no shuttlecock trajectories were assigned, the robot was rewarded for standing with the default joint configuration. This encouraged improved behavior during hardware deployment when the shuttle was not observed.

\begin{equation}
    r_{stand} = e^{-\frac{1}{\sigma N_{joints}}|q-q*|}
\end{equation}

\subsection*{Observations}

We provided the policy actor with observations that were available on the robot, and additional observations to the critic to help reduce value estimation error. The observations and their descriptions are listed in table~\ref{tab:observations}. Among the critic-only observations, those that provide complete \ac{MDP} information are highlighted in the grey cells.

\begin{table*}
    \centering
    \caption{Policy observations}
    \label{tab:observations}
    \begin{tabular}{p{1.8cm}p{3cm}p{12cm}}
    
    \cellcolor{grey}                                                   & \cellcolor{grey}\textbf{Observation}     & \cellcolor{grey}\textbf{Description}                                                                                                \\ \hline
    \multicolumn{1}{l|}{\multirow{12}{*}{shared}}      & $\mathbf{v}_b$                     & Base linear velocity                                                                                                \\
    \multicolumn{1}{l|}{}                              & $\mathbf{\omega}_b$ & Base angular velocity                                                                                               \\
    \multicolumn{1}{l|}{}                              & $\mathbf{g}_b$                     & Gravity vector in base frame                                                                                        \\
    \multicolumn{1}{l|}{}                              & $\mathbf{h}$                        & Robot heading vector in court frame                                                                                 \\
    \multicolumn{1}{l|}{}                              & cmd                      & Racket intercpetion command                                                                                         \\
    \multicolumn{1}{l|}{}                              & $t$                        & Time until the interception                                                                                         \\
    \multicolumn{1}{l|}{}                              & $\mathbf{q}$                        & Joint position offset from default configuration                                                                    \\
    \multicolumn{1}{l|}{}                              & $\mathbf{\Dot{q}}$                       & Joint velocity                                                                                                      \\
    \multicolumn{1}{l|}{}                              & $\mathbf{a}_{prev}$                  & Previous policy action output                                                                                       \\
    \multicolumn{1}{l|}{}                              & $\mathbf{p}_{B}^{court}$                      & Robot base position (x,y) in the court frame                                                                        \\
    \multicolumn{1}{l|}{}                              &   $\mathbf{p}_{shuttle}^{cam}$                      & Shuttlecock position in camera frame                                                                                \\
    \multicolumn{1}{l|}{}                              & $\mathcal{D}$         & Detected the shuttle in this timestep                                                                               \\ \hline
    \multicolumn{1}{l|}{\multirow{12}{*}{critic-only}} & \cellcolor{grey}$t_{hidden}$                & Time until the shuttlecock is launched                                                                              \\
    \multicolumn{1}{l|}{}                              & \cellcolor{grey}$t_{hidden_next}$          & Time until the next shuttle is launched                                                                             \\
    \multicolumn{1}{l|}{}                              & $\mathbf{p}_{EE}^{court}$                    & Racket sweet-spot position in the court frame                                                                       \\
    \multicolumn{1}{l|}{}                              & $\mathbf{v}^{court}_{ee}$                  & Racket sweet-spot velocity in the court frame                                                                       \\
    \multicolumn{1}{l|}{}                              & $\mathbf{n}_{x,racket}^{court}$            & Racket facing direction                                                                                             \\
    \multicolumn{1}{l|}{}                              & \cellcolor{grey}cmd$_{next}$                & The next interception position, velocity and orientation target                                                     \\
    \multicolumn{1}{l|}{}                              & \cellcolor{grey}Targets remaining      & Number of interception targets remaining in the episode                                                             \\
    \multicolumn{1}{l|}{}                              & \cellcolor{grey}$\mathcal{T}$          & Whether the robot is commanded to intercept a target or stand still                                                 \\
    \multicolumn{1}{l|}{}                              & \cellcolor{grey}$\mathcal{T}$ next     & Whether the robot is commanded to intercept a target or stand still after the current hit                           \\
    \multicolumn{1}{l|}{}                              & Domain randomization     & Randomly added base mass, robot friction coefficient                                                                \\
    \multicolumn{1}{l|}{}                              & $\epsilon$         & The difference between the target interception position and the one predicted from the \ac{EKF} shuttle state estimation \\
    \multicolumn{1}{l|}{}                              & Is in FOV                & Whether the shuttlecock is in the camera FOV                                                                       
    \end{tabular}
    \end{table*}

\subsection*{Training Hyperparameters}

The remaining training hyperparameters are presented in table \ref{tab:training_hyperparams}.

% Please add the following required packages to your document preamble:
% \usepackage[table,xcdraw]{xcolor}
% Beamer presentation requires \usepackage{colortbl} instead of \usepackage[table,xcdraw]{xcolor}

\begin{table}
    \centering
    \caption{Training Hyperparameters}
    \begin{tabular}{cc}
    \rowcolor[HTML]{EFEFEF} 
    \textbf{Parameter}          & \textbf{Value}  \\ \hline
    discount factor             & 0.995           \\
    GAE lambda                  & 0.975           \\
    learning rate               & adaptive        \\
    KLD target                  & 0.01            \\
    entropy coefficient         & 0.0016          \\
    entropy coeff. decay        & 0.99993         \\
    num. targets per episode    & 6               \\
    control dt (s)              & 0.01            \\
    terrain max height diff (m) & 0.06            \\
    num. envs                   & 4096            \\
    standing env. ratio         & 10\%            \\
    actor MLP size              & (512, 256, 128) \\
    critic MLP size             & (512, 256, 128) \\
    network activation          & elu             \\
    optimizer                   & AdamW          
    \end{tabular}
    \label{tab:training_hyperparams}
    \end{table}

\subsection*{State-dependent Action Standard Deviation}

\begin{figure}
    \centering
    \includegraphics[width=0.8\linewidth]{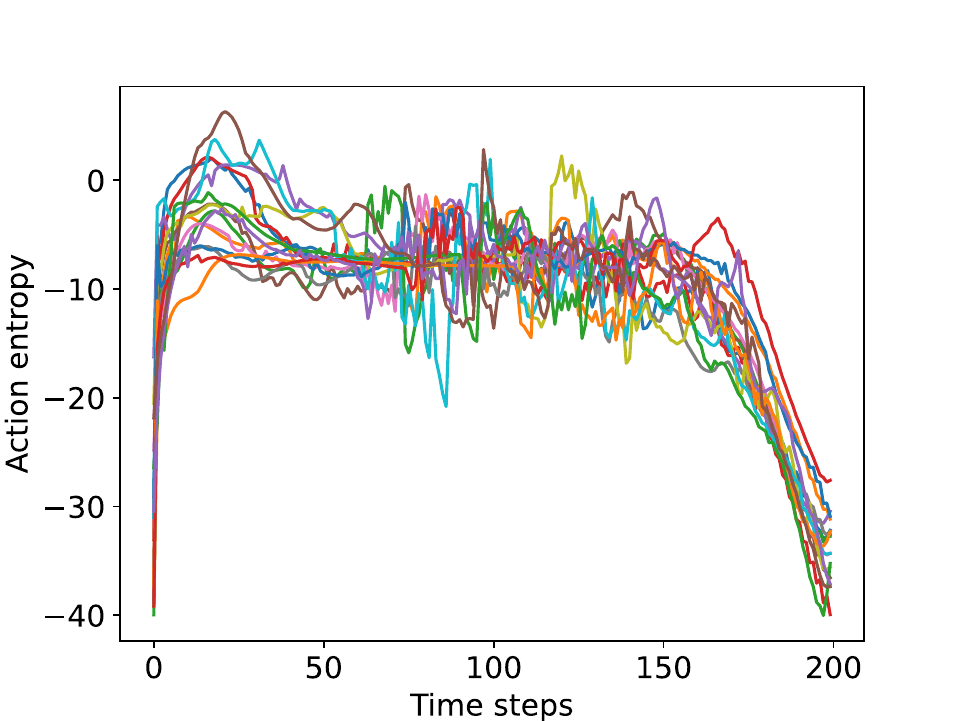}
    \caption{The policy's action distribution entropy decreases as the time approaches the swing at timestep=200. Plotted with 16 independent swings.}
    \label{fig:state_dependent_std}
\end{figure}

We used a state-dependent action standard deviation \cite{andrychowicz2020matters} for \ac{N-P3O}. Our evaluation shows that the action distribution entropy decreased in the timesteps leading up to the swing (fig.~\ref{fig:state_dependent_std}), resulting in a minor reward improvement during training.

\subsection*{Approximating the Estimated Interception}

During training, we estimated the shuttlecock interception position by linearizing the shuttle trajectory prediction~\cite{cohen2015physics} with respect to the state estimation error at the ground truth shuttlecock state. This simplification reduced the computational cost of full trajectory prediction based on shuttlecock state estimation. The process is outlined in Eq.~\ref{eq:badminton_pred_approx}.

\begin{equation}
    \hat{p}_T \approx p_T + (\hat{p}_{t_k} - p_{t_k}) + (1- \frac{2\Delta t}{L})(\hat{v}_{t_k}-v_{t_k}) (T-t_k) \label{eq:badminton_pred_approx}
\end{equation}

We rejected higher order $\Delta t$ term due to their small magnitudes.

\subsection*{Deflection Error}

The deflection error was calculated based on the difference in outgoing shuttlecock velocity between the expected result of the commanded racket swing (orientation and velocity) and the actual result from the executed swing. For this computation, we assumed a fixed nominal incoming shuttlecock velocity of $(v_x, v_y, v_z) = (-4.5, 0.0, -4.5)$.

We computed the velocity difference between the incoming shuttlecock and the commanded racket velocity as follows:

\begin{equation}
    \mathbf{v}_{\text{impact}} = \mathbf{v}_{\text{incoming}} - \mathbf{v}_{\text{target}}
\end{equation}

Next, we calculated the outgoing shuttlecock velocity, assuming an elastic collision where the reflected racket inertia was much larger than that of the shuttlecock:

\begin{align*}
    \mathbf{v}_{\text{racket\_normal}} & = (\mathbf{v}_{\text{racket}} \cdot \mathbf{n}_{\text{racket}}) \mathbf{n}_{\text{racket}}\\
    \mathbf{v}_{\text{shuttle\_normal}} & = (\mathbf{v}_{\text{impact}} \cdot \mathbf{n}_{\text{racket}}) \mathbf{n}_{\text{racket}}\\
    \mathbf{v}_{\text{outgoing}} & = \mathbf{v}_{\text{incoming}} - 2 \mathbf{v}_{\text{shuttle\_normal}} + 2 \mathbf{v}_{\text{racket\_normal}} 
\end{align*}

The same computation was applied to the executed swing, substituting the measured racket orientation and velocity to calculate the difference in the shuttlecock's outgoing velocity.

\subsection*{Perception Reaction Time}

\begin{figure}
    \centering
    \includegraphics[width=0.8\linewidth]{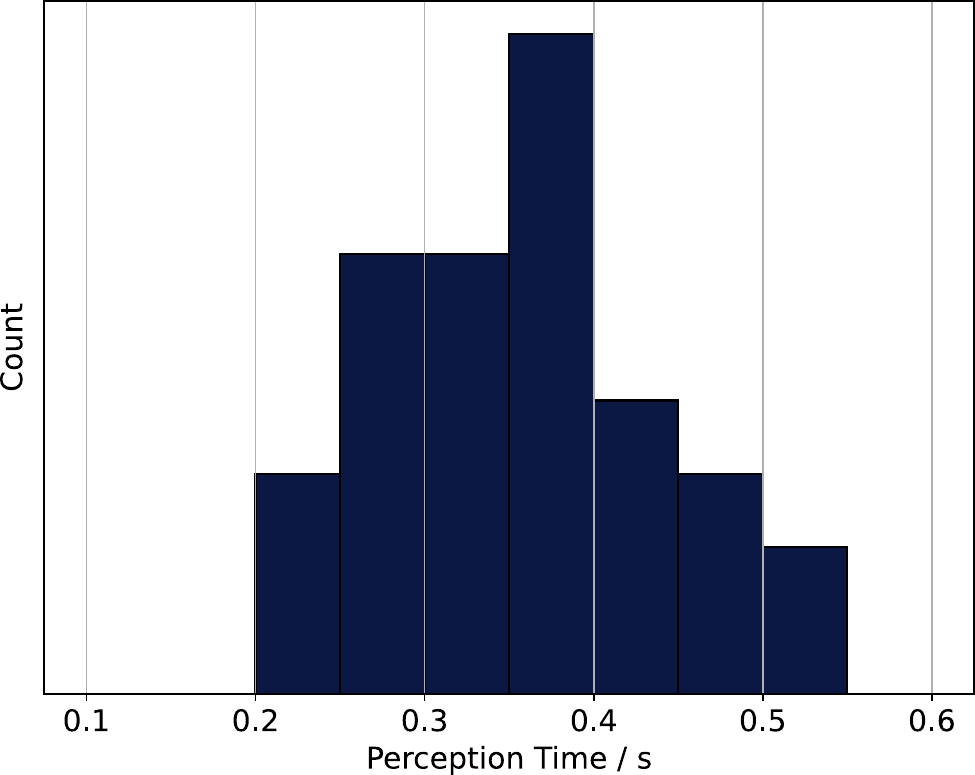}
    \caption{Distribution of the time between a human serving the shuttlecock and the completion of our system's perception loop to generate an \ac{EE} target.}
    \label{fig:perception_duration_hist}
\end{figure}

As noted in the main manuscript, the mean duration for our system to complete the perception loop and produce an \ac{EE} target was \unit[0.375]{s}. The distribution of this duration is shown in the histogram in fig.~\ref{fig:perception_duration_hist}, with a minimum duration of \unit[0.217]{s} and a maximum of \unit[0.517]{s}.

\subsection*{Additional Gait Adaptation}

\begin{figure}
    \centering
    \includegraphics[height=0.8\textheight]{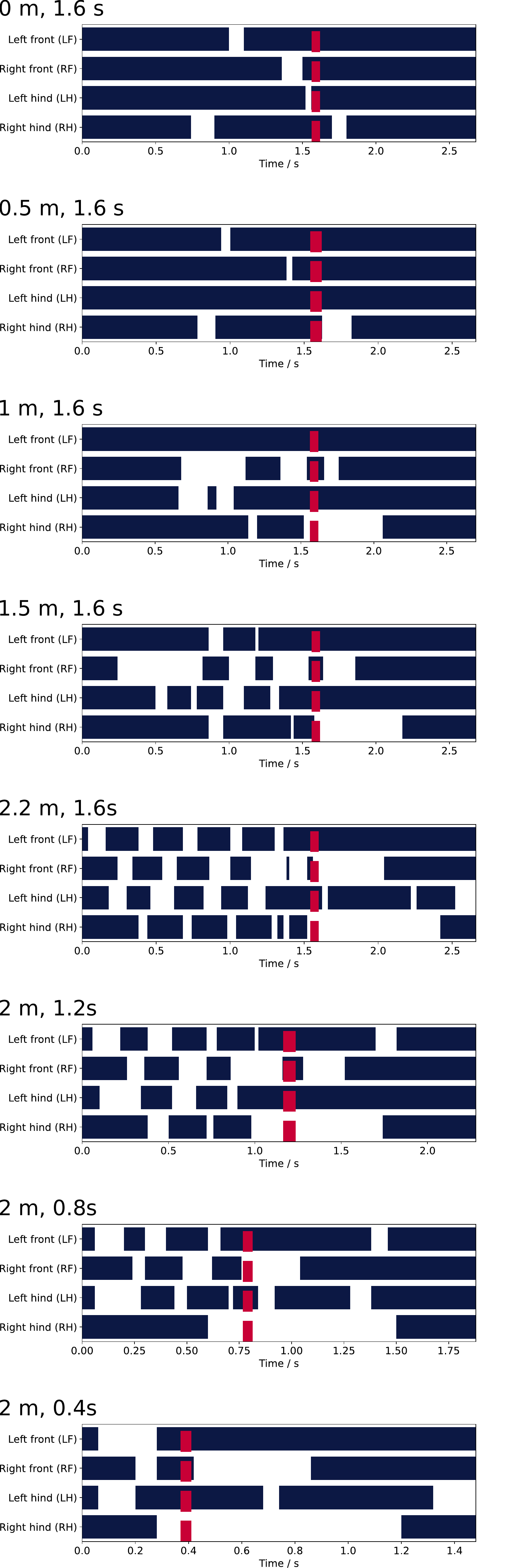}
    \caption{Foot contact schedule adaptation to various distances and duration.}
    \label{fig:full_gait}
\end{figure}

Due to space constraints, only a portion of the gait adaptation plot is shown in the main manuscript. The complete plots are provided in fig.~\ref{fig:full_gait}.

\subsection*{Potential Extensions to Other Robots}

We applied our training framework to humanoid robots (Unitree G1) in simulation for badminton. The humanoids demonstrated agile, coordinated whole-body visuomotor skills comparable to ANYmal, as shown in movie S5. We acknowledge that deploying this on hardware presents notable challenges, and this experiment serves only as an early indication of potential future directions. Table~\ref{tab:humanoid_config} summarizes the adjusted training configurations; all other settings remained unchanged.

\begin{table}
    \centering
    \caption{Humanoid Training Configurations}
    \begin{tabular}{cc}
    \rowcolor[HTML]{EFEFEF} 
    \textbf{Config}                     & \textbf{Changes}            \\ \hline
    Robot                                      & Unitree G1      \\
    Num. DoFs                                  & 23                          \\
    Current limit & Removed                     \\
    Motor model                                & From \text{unitree\_rl\_gym}             
    \end{tabular}
    \label{tab:humanoid_config}
    \end{table}

\subsection*{Alternative Joint Regularization Settings}

\begin{figure}
    \centering
    \includegraphics[width=0.9\linewidth]{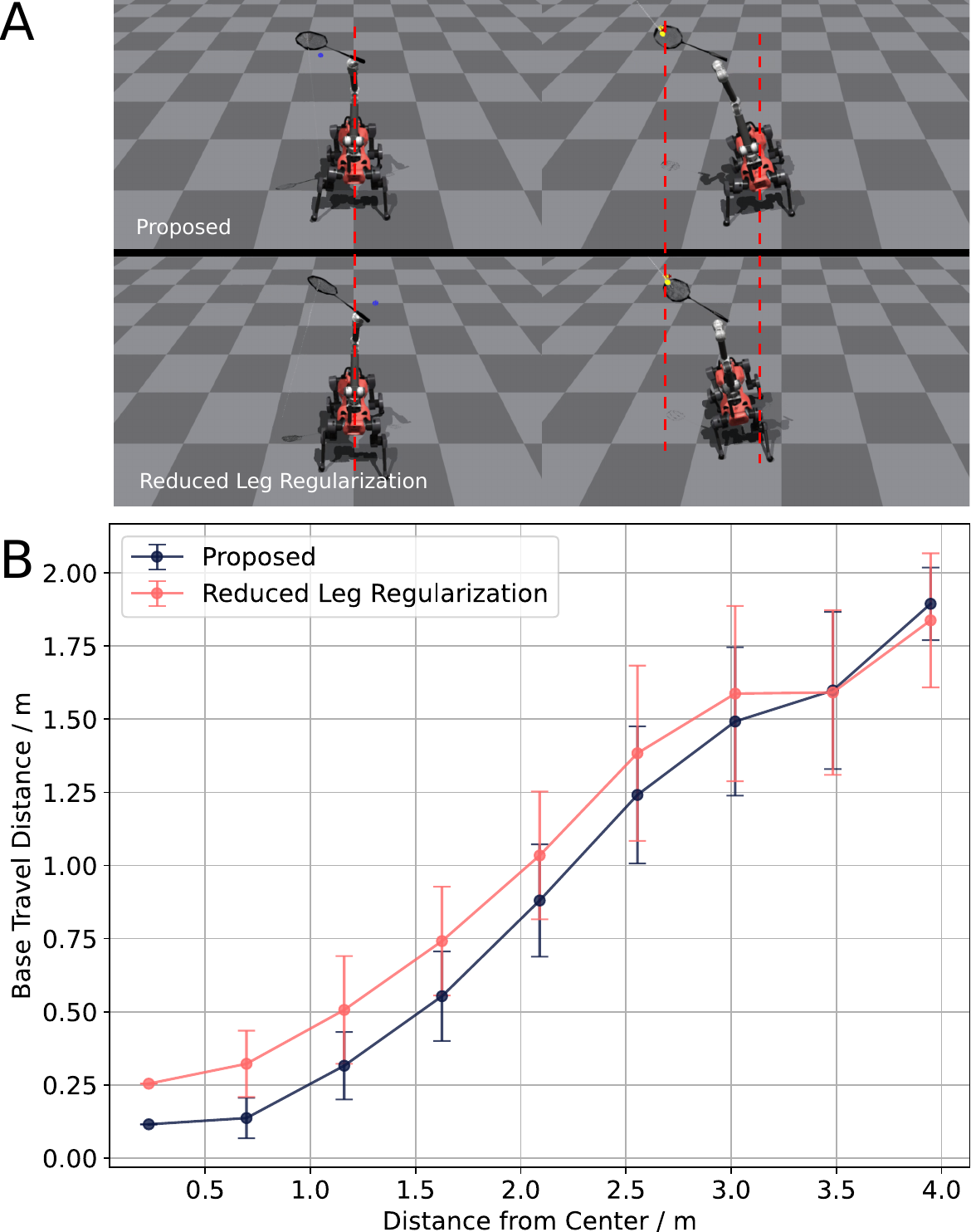}
    \caption{A comparison of base displacement for various target distances between the default training rewards and reduced leg regularization. \textbf{(A)} Different policies lead to distinct base travel when hitting the same target from the same initial position. \textbf{(B)} The policy with less regularized leg motion shows larger base displacement when hitting nearby targets compared to the default weighting. The line plots and error bars show the mean and standard deviation of the base displacement in 4096 test trajectories.}
    \label{fig:base_travel_dist}
\end{figure}

We observed that the robot prioritized different joints if the joint torque and acceleration regularization rewards were scaled differently, resulting in distinct behaviors. As shown in Figure~\ref{fig:base_travel_dist}, when the policy was trained with reduced leg joint regularization, the robot exhibited greater base displacements for nearby targets, though distant targets that inherently required dynamic leg motion remained largely unaffected. Both regularization terms were reduced to 1/3 of their original values for the reduced regularization setting.

\clearpage

%%%%%%%%%%%%%%%% SUPPLEMENTARY FIGURES %%%%%%%%%%%%%%%

\end{document}